\documentclass[11pt]{article}
\usepackage[final]{acl}
\usepackage{times}
\usepackage{latexsym}
\usepackage[T1]{fontenc}
\usepackage[utf8]{inputenc}
\usepackage{microtype}
\usepackage{inconsolata}
\usepackage{graphicx}
\usepackage{tikz}
\usetikzlibrary{decorations.pathreplacing}
\usepackage{amsmath}
\usepackage{amssymb}
\usepackage{booktabs}
\usepackage{multirow}
\usepackage{algorithm}
\usepackage{algorithmic}
\usepackage{xcolor}
\usepackage{subcaption}
\usepackage[disable]{todonotes}

\definecolor{my_yellow}{RGB}{248,236,198}
\definecolor{my_green}{RGB}{197,224,180}

\title{Semantic Head Specialization Guides Hybrid ViT Attention\\for Multimodal LLMs}
\author{
Chenhong He\textsuperscript{1,3}, Lei Li\textsuperscript{2,3}, Shicheng Li\textsuperscript{3}, Hanglong Lv\textsuperscript{1,3} \\
\bfseries Lingpeng Kong\textsuperscript{2},
Qi Liu\textsuperscript{2},
Tong Yang\textsuperscript{1,\textdagger},
Shuhuai Ren\textsuperscript{3,\textdagger} \\
\textsuperscript{1}State Key Laboratory of Multimedia Information Processing, \\
School of Computer Science, Peking University \\
\textsuperscript{2}Department of Computer Science, The University of Hong Kong \\
\textsuperscript{3}Xiaomi Corporation, LLM-Core Team \\
\texttt{heccc@stu.pku.edu.cn, nlp.lilei@gmail.com} \\
\textsuperscript{\textdagger}Corresponding authors.
}
\begin{document}
\maketitle


\begin{abstract}
Hybrid attention dominates frontier LLMs, yet Vision Transformers (ViTs) in multimodal LLMs lack a satisfactory hybrid design, with no consensus on why certain attention patterns work better.
To fill this gap, we study ViT attention heads and find they differentiate into object- and background-specialist roles, a pattern most pronounced under full attention; we call this Semantic Head Specialization (SHS).
We propose SHS-Index to quantify this specialization, show that it distinguishes full-attention from chunk-window ViTs, and find that it correlates with downstream benchmark performance within our controlled design space.
We then identify three structural factors that shape SHS---window interaction, token serialization, and local softmax allocation---and use them as design principles for hybrid attention.
Guided by these factors, we design Ariadne Attention, a hybrid within about 0.5 points of full attention on the 20-image benchmark at $6.5\times$ lower attention FLOPs; its end-to-end ViT time is 13.5\% lower at $896^2$.
Our findings establish head specialization as a measurable property for diagnosing and designing principled hybrid ViT attention at the multimodal-LLM scale.
\end{abstract}

\section{Introduction}
\label{sec:intro}

\begin{figure}[!t]
\centering
\includegraphics[width=\columnwidth]{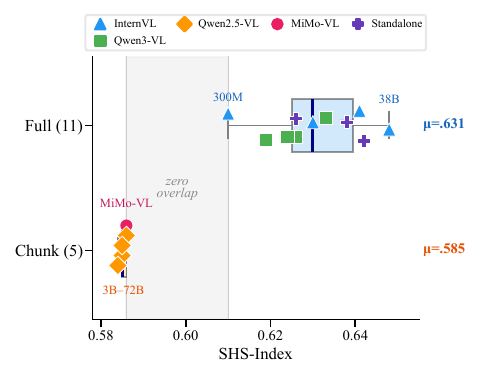}
\caption{\textbf{SHS-Index separates ViT attention designs.} Across 16 open-source visual encoders and VLMs, SHS-Index cleanly separates full-attention and chunk-window ViTs. Within Qwen2.5-VL, changing the LLM size from 3B to 72B leaves SHS-Index nearly unchanged, indicating that the signal comes from the ViT attention design.}
\label{fig:hero_a}
\end{figure}

\begin{figure*}[t]
\centering
\setlength{\tabcolsep}{1.5pt}
\renewcommand{\arraystretch}{0.85}
\begin{tabular}{@{}c ccc c@{}}
\small\textbf{Input} & \small\textbf{Full Attention} & \small\textbf{Ariadne (Ours)} & \small\textbf{Chunk Window} & \\[2pt]
\includegraphics[width=0.215\textwidth]{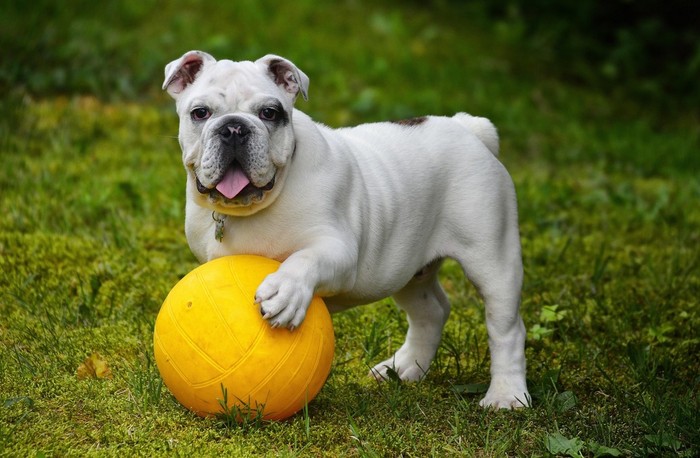} &
\includegraphics[width=0.215\textwidth]{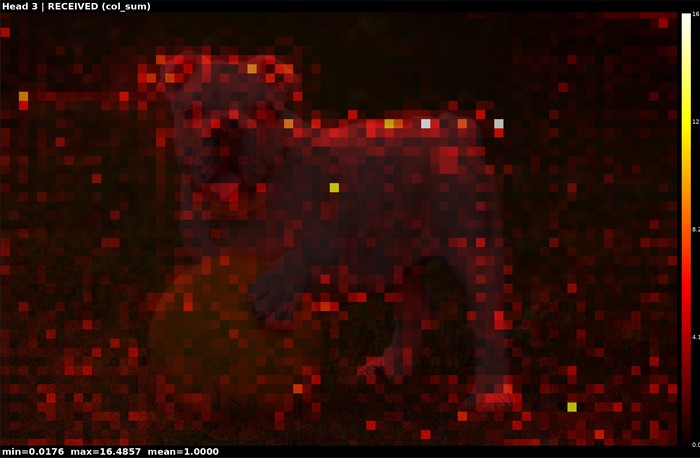} &
\includegraphics[width=0.215\textwidth]{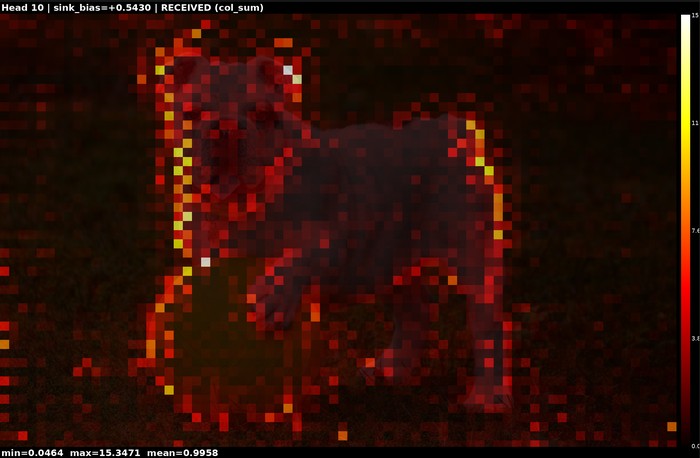} &
\includegraphics[width=0.215\textwidth]{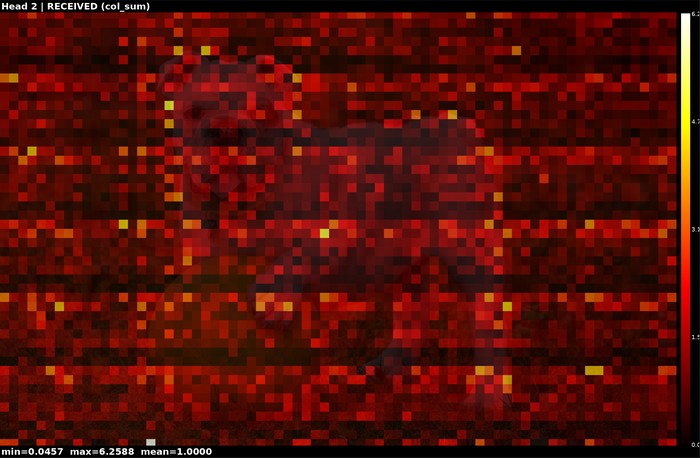} &
\small\;\;L11 \\[1.5pt]
\includegraphics[width=0.215\textwidth]{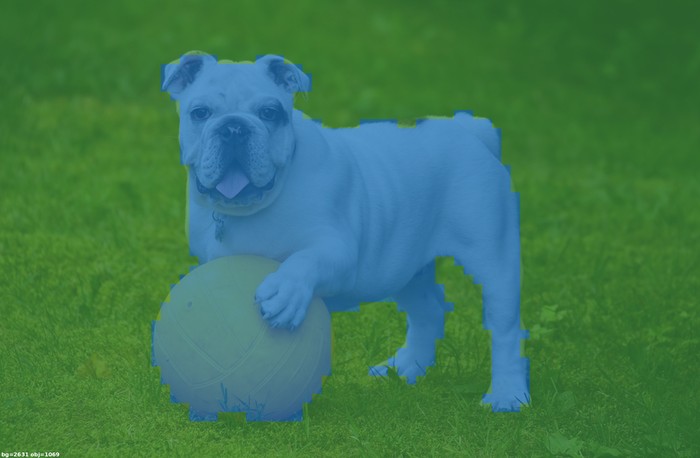} &
\includegraphics[width=0.215\textwidth]{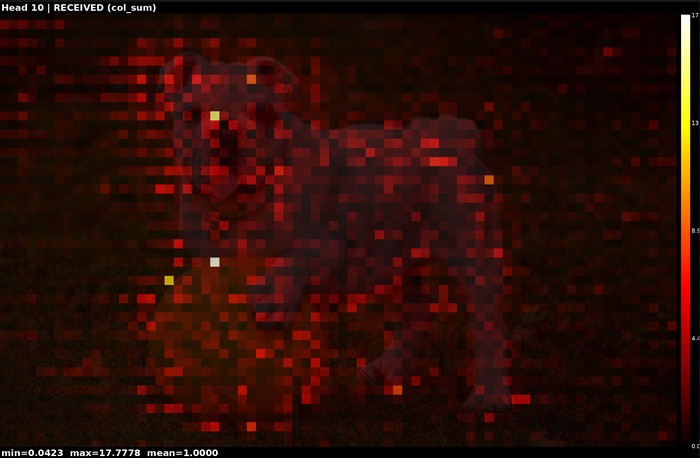} &
\includegraphics[width=0.215\textwidth]{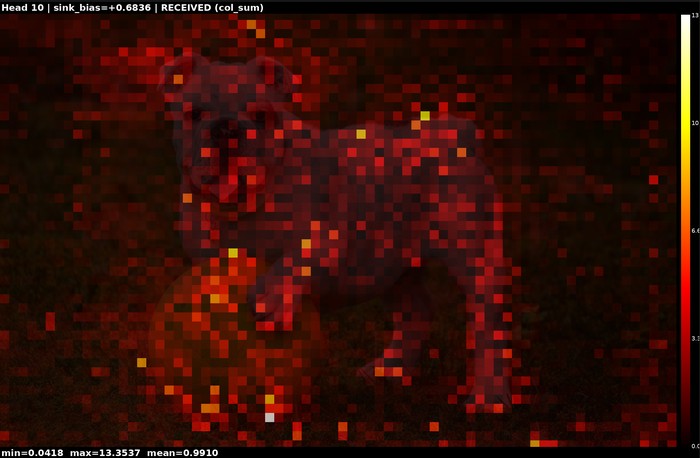} &
\includegraphics[width=0.215\textwidth]{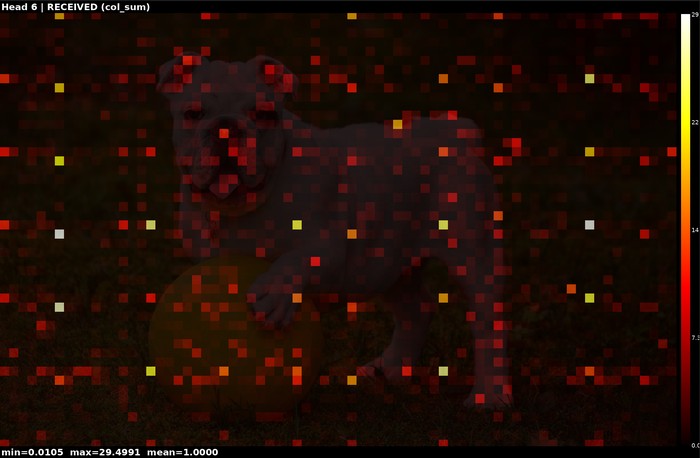} &
\small\;\;L17 \\
\end{tabular}
\caption{\textbf{Semantic Head Specialization (SHS) in ViT attention heads.} \emph{First column:} input image (top) and a foreground (blue) / background (green) segmentation (bottom); each ViT token is labeled foreground or background by the dominant region within its image patch. \emph{Columns 2--4:} received-attention heatmaps for a representative head at two ViT layers (L11, L17) across three attention designs. Under Full Attention, heads specialize: bright regions concentrate on the foreground object, revealing clear object/background differentiation. Our Ariadne attention preserves this specialization. Under Chunk Window attention (Qwen2.5-VL style), heads show grid-like window-boundary artifacts and fail to differentiate foreground from background.
}
\label{fig:hero}
\end{figure*}

As Vision Transformers (ViTs)~\citep{dosovitskiy2021vit} scale to higher resolutions, full self-attention becomes increasingly expensive, making efficient attention a central design problem.
Hybrid attention is a natural candidate: frontier LLMs such as GPT-OSS \citep{openai2025gptoss} and Gemma~3 \citep{gemmateam2025gemma3} reduce cost by interleaving local and full-attention layers.
In ViTs, however, existing hybrids have not yet matched the quality of full attention.
Widely used efficient ViT designs such as Qwen2.5-VL \citep{bai2025qwen25vl} rely on chunk-window locality, where tokens are grouped into non-overlapping windows, while recent high-quality VLM encoders such as Qwen3-VL \citep{bai2025qwen3vl} return to full attention.
This quality gap raises a concrete question: what does full attention preserve that current chunk-window ViT hybrids lose?

We study this question with a controlled full-vs.-chunk comparison, choosing chunk windows because they are the main deployed ViT hybrid baseline in current VLMs.
We train two ViTs from scratch under identical conditions, ensuring that the attention operator is the only difference: one model uses full attention, while the other uses Qwen-style chunk windows.
We then inspect per-head attention maps to see how the two operators organize visual information internally.
The maps reveal a clear qualitative contrast: under full attention, heads differentiate into semantic roles, with some focusing on foreground objects and others on the background; under chunk windows, heads instead show more uniform grid-like patterns.
We name this phenomenon \textbf{Semantic Head Specialization (SHS)}.
To make this observation measurable beyond individual heatmaps, we propose \textbf{SHS-Index}, an AUROC-based metric of how strongly each head separates foreground from background tokens.
Across 16 open-source visual encoders and VLMs, SHS-Index separates full-attention and chunk-window ViTs with no overlap in score ranges, showing that the full-vs.-chunk specialization gap generalizes beyond our controlled training pair.
Within Qwen2.5-VL, changing only the LLM backbone size from 3B to 72B changes SHS-Index by less than $0.002$, supporting that the signal is tied to ViT attention design rather than LLM scale (Figure~\ref{fig:hero_a}).
These results justify using SHS-Index as a ViT-side diagnostic for attention design.

We next use SHS-Index to diagnose why chunk-window attention weakens specialization and how a better hybrid can recover it.
We focus on three structural factors---window interaction, token serialization, and local softmax allocation---and measure how each one changes SHS-Index.
After this structural analysis, we evaluate the sufficiently trained diagnostic configurations on 22 downstream benchmarks.
At the primary 8k checkpoint, SHS-Index correlates with the corrected 22-task average (Pearson $r{=}0.858$, $p{=}0.006$; exact permutation $p{=}0.008$), with the relationship interpreted as cross-architecture evidence rather than a causal or within-run predictor.
Guided by the validated factors, we design \textbf{Ariadne Attention}\footnote{Named after Ariadne's thread in Greek mythology: as the thread guided Theseus through the labyrinth, sliding-window attention traces a continuous path through the token sequence.}, an SHS-informed hybrid combining sliding windows, per-head sink bias, and row-column alternating serialization.
Ariadne reaches 40.40 versus 40.92 for full attention on the 20-image benchmark, while reducing attention FLOPs by $6.5\times$ and end-to-end ViT time by 13.5\% at $896^2$ (Appendix~\ref{sec:appx_efficiency}).
These results establish head specialization as a measurable property linking attention design to ViT quality, opening a path toward principled hybrid ViT attention design.

Our contributions are:
\textbf{(1)} We discover Semantic Head Specialization (SHS), the phenomenon that ViT heads differentiate into object- and background-specialist roles, and propose SHS-Index, a diagnostic metric that separates attention families across 16 open-source visual encoders and VLMs.
\textbf{(2)} We identify three architectural factors that shape SHS---window interaction, token serialization, and local softmax allocation---through targeted structural comparisons.
\textbf{(3)} We show that SHS-Index correlates with benchmark performance across controlled architectures at the primary checkpoint, and use it to design Ariadne Attention, which comes within about 0.5 points of full attention on 20 image tasks at $6.5\times$ lower attention FLOPs.

\section{Semantic Head Specialization in ViT Attention}
\label{sec:shs}

To compare how full-attention and chunk-window ViTs differ in attention behavior and downstream quality, we first study their per-head attention patterns under a controlled full-vs.-chunk comparison (\S\ref{sec:setup}).
We quantify this phenomenon with \textbf{SHS-Index} (\S\ref{sec:shs_index}) and analyze whether the specialization gap is an architecture-level property rather than merely a side effect of LLM scale.

\subsection{Full-vs.-Chunk Comparison of Attention Patterns}
\label{sec:setup}
\label{sec:discovery}

We aim to understand how different ViT attention operators shape per-head attention patterns.
Rather than starting from downstream scores, we inspect the attention maps induced by two specific operators: \textbf{full attention}, where every token can attend to every other token, and \textbf{chunk-window attention}, where non-overlapping chunks restrict each token to attend only within its own chunk.
To compare these operators, we start from the attention computation itself.
For head $h$ over a sequence of $N$ visual tokens, let $\mathcal{N}(i)$ be the tokens that query $i$ attends to, and define $\mathcal{I}(a,b)=\{a,a+1,\ldots,b\}$ as the set of token indices from $a$ to $b$.
For full attention, $\mathcal{N}(i)=\mathcal{I}(1,N)$.
For chunk attention with chunk size $C$ (e.g., $\ll N$), $\mathcal{N}(i)=\mathcal{I}(b_iC{+}1,\min(N,(b_i{+}1)C))$, where $b_i=\lfloor(i-1)/C\rfloor$.
The head-specific attention weight from query $i$ to key $j$ is:
\begin{equation}
\label{eq:attn_weight}
\small
a^{(h)}_{ij} = \frac{\exp(\mathbf{q}_i^\top \mathbf{k}_j / \sqrt{d})}{\sum_{j' \in \mathcal{N}(i)} \exp(\mathbf{q}_i^\top \mathbf{k}_{j'} / \sqrt{d}) + \exp(s^{(h)}_{\mathrm{aux}})}
\end{equation}
where $\mathbf{q}_i$ and $\mathbf{k}_j$ are the query and key vectors for tokens $i$ and $j$, $d$ is the per-head dimension, and $s^{(h)}_{\mathrm{aux}}$ is a per-head learnable sink bias \citep{xiao2024streamingllm, darcet2024vision} (set to $-\infty$ when absent).
We use \emph{received attention} as each token's importance measure: the total attention mass that token collects from all queries that can see it.
Received attention for token $j$ is:
\begin{equation}
\label{eq:rcv}
r_h(j) = \sum_{i:\, j \in \mathcal{N}(i)} a^{(h)}_{ij}
\end{equation}
Intuitively, $r_h(j)$ is token $j$'s head-$h$ importance: total votes from queries that can see it.

A fair comparison must not conflate attention type with layer depth or training conditions.
We therefore train a matched pair of 32-layer ViT encoders.
One model uses full self-attention in all layers.
The other follows Qwen2.5-VL-style: layers 7, 15, 23, and 31 use full attention, and the remaining 28 layers replace full attention with chunk-window attention \citep{bai2025qwen25vl} (architecture details in Appendix~\ref{sec:appx_vit_arch}).
We then visualize per-head received-attention maps on COCO val2017 images \citep{lin2014microsoft}. The results reveal a striking difference between the two models.
Under full attention (Figure~\ref{fig:hero}, column 2), heads do not merely produce different-looking maps; they organize into consistent semantic roles.
Some heads assign high received attention to foreground objects, with bright regions aligned with the dog and ball, while others assign high received attention to background regions.
We call this phenomenon \textbf{Semantic Head Specialization (SHS)}: attention heads specialize by separating semantic regions of the image, here most clearly visible as distinct object- and background-oriented roles.

The chunk-window baseline shows much weaker SHS (Figure~\ref{fig:hero}, column 4).
Its heatmaps exhibit grid-like artifacts from window boundaries, and different heads look qualitatively similar rather than forming clear semantic roles.
This contrast suggests that the quality gap between full and chunk-window attention is reflected not only in downstream benchmark scores, but also in how attention heads organize internally and assign each token to a foreground- or background-oriented role.

SHS is also not a single-image artifact.
The same head maintains a consistent foreground-or-background preference across all analysis images (per-image heatmaps in Appendix~\ref{sec:appx_heatmaps}; iteration-aggregated object dominance in Appendix~\ref{sec:appx_shs_data}), indicating that SHS is a stable property of the trained model rather than a stimulus-dependent visualization effect.
The pattern further generalizes across diverse visual scenes---including single objects, multiple objects, complex backgrounds, and scenes with overlapping or cluttered foreground entities---and consistently maintains object/background head specialization (Appendix~\ref{sec:appx_heatmaps}).

\subsection{Quantifying Head Specialization}
\label{sec:shs_index}

The heatmaps above suggest that a specialized head should consistently prefer one semantic side of the image: either object tokens or background tokens.
We therefore turn SHS into a foreground--background separation problem.
For this quantitative analysis, we use COCO segmentation masks to assign each visual token to foreground or background according to the dominant region in its image patch (details in Appendix~\ref{sec:appx_setup}).
Given the received-attention score $r_h(j)$ (Eq.~\ref{eq:rcv}), a foreground-specialist head should give larger scores to foreground tokens than to background tokens, while a background-specialist head should consistently do the opposite.

For each head $h$ and image, we compute the AUROC of received attention $r_h(j)$ for ranking foreground tokens above background tokens:
\begin{equation}
\begin{aligned}
\text{AUROC}_h ={}& \Pr(r_h(j^+) > r_h(j^-)) \\
& + \tfrac{1}{2}\Pr(r_h(j^+) = r_h(j^-)),
\end{aligned}
\end{equation}
where $j^+$ and $j^-$ denote foreground and background tokens.
AUROC values near $1$ or $0$ indicate foreground- or background-oriented heads, while values near $0.5$ indicate no separation.
Since either direction (FG/BG) indicates specialization, we define the head-level score as
\begin{equation}
\text{SHS}_h=\max(\text{AUROC}_h,\,1-\text{AUROC}_h).
\end{equation}
which ranges from $0.5$ (no separation) to $1$ (strong specialization).
The final \textbf{SHS-Index} averages this score over images, layers, and heads:
\begin{equation}
\text{SHS-Index} = \mathbb{E}_{\mathrm{img},\ell,h}[\text{SHS}_h].
\end{equation}

On our controlled full-vs.-chunk pair, SHS-Index is .606 for Full Attention and .577 for Chunk Window, quantitatively matching the specialization gap observed in the heatmaps.

\paragraph{SHS-Index separates attention families in open-source models.}
The controlled pair above shows that SHS-Index captures the specialization gap between full and chunk-window attention under matched training conditions.
We next ask whether the same signal appears in existing VLMs rather than only in our controlled models.
Across 16 open-source visual encoders and VLMs, SHS-Index cleanly separates the 11 full-attention encoders (mean SHS\,$\approx$\,0.631) from the 5 chunk-window ViTs (mean SHS\,$=$\,0.585), with no observed overlap (Figure~\ref{fig:hero_a}; full data in Appendix~\ref{sec:appx_shs_data}).
Moreover, the identical Qwen2.5-VL ViT changes by only ${<}\,0.002$ when paired with 3B--72B LLM backbones, showing that SHS-Index is determined by the ViT attention design rather than by the language-model scale.
Thus, the open-source analysis supports SHS as an architecture-level diagnostic. We do not use these heterogeneous models for benchmark correlation because their data, training recipes, and language backbones differ.

\begin{figure*}[!t]
\centering
\begin{tikzpicture}[font=\scriptsize]
\def\s{0.58}

\node[font=\footnotesize\bfseries, anchor=south] at ({4*\s}, 0.3) {(a) Raster-scan};

\foreach \row in {0,...,3}{
  \foreach \col in {0,...,7}{
    \pgfmathtruncatemacro{\val}{\row*8+\col}
    \pgfmathtruncatemacro{\bparity}{mod(int(\row/2)+int(\col/2),2)}
    \ifnum\bparity=0
      \fill[blue!10] ({\col*\s},{-(\row+1)*\s}) rectangle +(\s,\s);
    \else
      \fill[orange!12] ({\col*\s},{-(\row+1)*\s}) rectangle +(\s,\s);
    \fi
    \draw[thin] ({\col*\s},{-(\row+1)*\s}) rectangle +(\s,\s);
    \node at ({(\col+0.5)*\s},{-(\row+0.5)*\s}) {\val};
  }
}
\foreach \bi in {0,1}{
  \foreach \bj in {0,...,3}{
    \draw[thick] ({\bj*2*\s},{-\bi*2*\s}) rectangle +({2*\s},{-2*\s});
  }
}

\begin{scope}[xshift=6.2cm]
\node[font=\footnotesize\bfseries, anchor=south] at ({4*\s}, 0.3) {(b) Chunk window (Qwen2.5-VL)};

\foreach \row/\col/\val in {
  0/0/0,  0/1/1,  0/2/4,  0/3/5,  0/4/16, 0/5/17, 0/6/20, 0/7/21,
  1/0/2,  1/1/3,  1/2/6,  1/3/7,  1/4/18, 1/5/19, 1/6/22, 1/7/23,
  2/0/8,  2/1/9,  2/2/12, 2/3/13, 2/4/24, 2/5/25, 2/6/28, 2/7/29,
  3/0/10, 3/1/11, 3/2/14, 3/3/15, 3/4/26, 3/5/27, 3/6/30, 3/7/31
}{
    \pgfmathtruncatemacro{\bparity}{mod(int(\row/2)+int(\col/2),2)}
    \ifnum\bparity=0
      \fill[blue!10] ({\col*\s},{-(\row+1)*\s}) rectangle +(\s,\s);
    \else
      \fill[orange!12] ({\col*\s},{-(\row+1)*\s}) rectangle +(\s,\s);
    \fi
    \draw[thin] ({\col*\s},{-(\row+1)*\s}) rectangle +(\s,\s);
    \node at ({(\col+0.5)*\s},{-(\row+0.5)*\s}) {\val};
}
\foreach \bi in {0,1}{
  \foreach \bj in {0,...,3}{
    \draw[thick] ({\bj*2*\s},{-\bi*2*\s}) rectangle +({2*\s},{-2*\s});
  }
}
\draw[red!70!black, line width=1.8pt, dashed] ({4*\s}, 0) -- ({4*\s}, {-4*\s});
\node[font=\tiny, red!70!black, anchor=south] at ({2*\s}, 0.05) {Window 0};
\node[font=\tiny, red!70!black, anchor=south] at ({6*\s}, 0.05) {Window 1};
\end{scope}

\begin{scope}[yshift=-3.5cm]
\node[font=\footnotesize\bfseries, anchor=south] at ({4*\s}, 0.3) {(c) Row-major (Ours)};

\foreach \row/\col/\val in {
  0/0/0,  0/1/1,  0/2/4,  0/3/5,  0/4/8,  0/5/9,  0/6/12, 0/7/13,
  1/0/2,  1/1/3,  1/2/6,  1/3/7,  1/4/10, 1/5/11, 1/6/14, 1/7/15,
  2/0/16, 2/1/17, 2/2/20, 2/3/21, 2/4/24, 2/5/25, 2/6/28, 2/7/29,
  3/0/18, 3/1/19, 3/2/22, 3/3/23, 3/4/26, 3/5/27, 3/6/30, 3/7/31
}{
    \pgfmathtruncatemacro{\bparity}{mod(int(\row/2)+int(\col/2),2)}
    \ifnum\bparity=0
      \fill[blue!10] ({\col*\s},{-(\row+1)*\s}) rectangle +(\s,\s);
    \else
      \fill[orange!12] ({\col*\s},{-(\row+1)*\s}) rectangle +(\s,\s);
    \fi
    \draw[thin] ({\col*\s},{-(\row+1)*\s}) rectangle +(\s,\s);
    \node at ({(\col+0.5)*\s},{-(\row+0.5)*\s}) {\val};
}
\foreach \bi in {0,1}{
  \foreach \bj in {0,...,3}{
    \draw[thick] ({\bj*2*\s},{-\bi*2*\s}) rectangle +({2*\s},{-2*\s});
  }
}
\end{scope}

\begin{scope}[xshift=6.2cm, yshift=-3.5cm]
\node[font=\footnotesize\bfseries, anchor=south] at ({4*\s}, 0.3) {(d) Column-major (Ours)};

\foreach \row/\col/\val in {
  0/0/0,  0/1/1,  0/2/8,  0/3/9,  0/4/16, 0/5/17, 0/6/24, 0/7/25,
  1/0/2,  1/1/3,  1/2/10, 1/3/11, 1/4/18, 1/5/19, 1/6/26, 1/7/27,
  2/0/4,  2/1/5,  2/2/12, 2/3/13, 2/4/20, 2/5/21, 2/6/28, 2/7/29,
  3/0/6,  3/1/7,  3/2/14, 3/3/15, 3/4/22, 3/5/23, 3/6/30, 3/7/31
}{
    \pgfmathtruncatemacro{\bparity}{mod(int(\row/2)+int(\col/2),2)}
    \ifnum\bparity=0
      \fill[blue!10] ({\col*\s},{-(\row+1)*\s}) rectangle +(\s,\s);
    \else
      \fill[orange!12] ({\col*\s},{-(\row+1)*\s}) rectangle +(\s,\s);
    \fi
    \draw[thin] ({\col*\s},{-(\row+1)*\s}) rectangle +(\s,\s);
    \node at ({(\col+0.5)*\s},{-(\row+0.5)*\s}) {\val};
}
\foreach \bi in {0,1}{
  \foreach \bj in {0,...,3}{
    \draw[thick] ({\bj*2*\s},{-\bi*2*\s}) rectangle +({2*\s},{-2*\s});
  }
}
\end{scope}

\end{tikzpicture}
\caption{Token arrangements on a $4\!\times\!8$ patch grid ($2\!\times\!4$ blocks). Numbers indicate position in the 1D sequence; bold borders delimit $2\!\times\!2$ blocks. \textbf{(a)}~Raster-scan. \textbf{(b)}~Qwen2.5-VL chunk window: blocks grouped into non-overlapping windows (dashed red line); token~15 and token~16 are spatially adjacent but in separate windows. \textbf{(c)}~Row-major: all blocks serialized left-to-right, top-to-bottom for SWA. \textbf{(d)}~Column-major: blocks serialized top-to-bottom, left-to-right. Ariadne alternates (c) and (d) across layers (\S\ref{sec:ariadne}).}
\label{fig:arrangement}
\end{figure*}
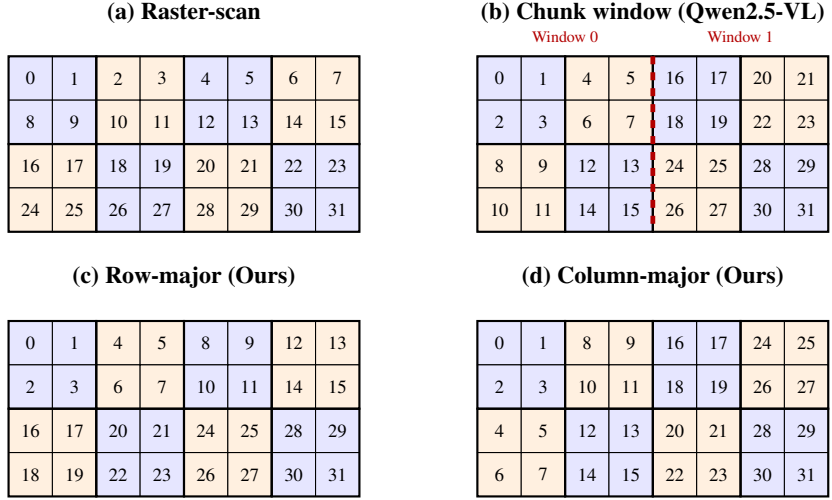

\section{Structural Sources of the SHS Gap}
\label{sec:findings}

The evidence above points to a clear SHS gap: full attention specializes heads much more strongly than chunk-window attention.
We now ask which structural choices create this gap, with the goal of turning them into design principles for a better hybrid attention.
We study three factors in turn---window interaction, token serialization, and local softmax allocation---using controlled attention-only comparisons.
As in \S\ref{sec:setup}, all comparisons keep the 32-layer ViT, data, training recipe, and full-attention anchor layers fixed; only the attention structure in the remaining 28 layers changes.

\paragraph{Factor 1: Window isolation restricts spatial scope.}
Chunk-window attention differs from full attention most directly in spatial reach.
In full attention, every token can collect evidence from the whole image.
In chunk-window attention, image tokens are partitioned into non-overlapping chunks, so a local layer can only mix information inside each chunk.
If SHS requires heads to compare object and background evidence across regions, then the first question is whether relaxing this chunk-level isolation improves SHS.
To answer this question, we keep Qwen2.5-VL's original 2D block layout, where image patches are grouped into local $2{\times}2$ blocks and then ordered as chunk windows (Figure~\ref{fig:arrangement}b), and change only the attention operator from chunk-window attention to sliding-window attention (SWA) \citep{jiang2023mistral}.
Here, SWA denotes a token-wise local attention pattern: with window $(l,r)$, query $i$ attends to $\mathcal{N}_{\mathrm{SWA}}(i)=\mathcal{I}(\max(1,i{-}l),\min(N,i{+}r))$, i.e., a window of preceding and following tokens rather than a fixed chunk.
Unlike fixed chunks, this window overlaps across chunk boundaries, so tokens are no longer confined to their original chunk; across layers, information can propagate through successive windows to much farther spatial regions.
Under this controlled change, SHS-Index increases from .577 (chunk-2D) to .588 (SWA-2D).
Attention distance provides a direct physical check: using the attention distance of \citet{dosovitskiy2021vit}, $D = \mathbb{E}_{h,i}[\sum_j a^{(h)}_{ij}\|p_i - p_j\|_2]$, where $p_i$ is the 2D image position of token $i$, the average span rises from $D=10.69$ to $D=11.80$.
\textbf{This identifies window isolation as one source of the SHS gap: chunk windows restrict spatial scope, and a better hybrid ViT should propagate spatial context by allowing token interaction to flow freely across window boundaries.}

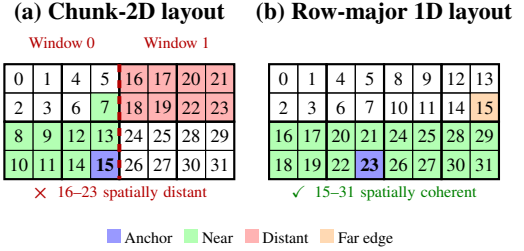
\begin{figure}[!t]
\centering
\begin{tikzpicture}[font=\scriptsize]
\def\s{0.38}

\node[font=\footnotesize\bfseries, anchor=south] at ({4*\s}, 0.4) {(a) Chunk-2D layout};

\foreach \row/\col/\val in {0/0/0, 0/1/1, 0/2/4, 0/3/5, 1/0/2, 1/1/3, 1/2/6, 2/4/24, 2/5/25, 2/6/28, 2/7/29, 3/4/26, 3/5/27, 3/6/30, 3/7/31}{
    \draw[thin] ({\col*\s},{-(\row+1)*\s}) rectangle +(\s,\s);
    \node at ({(\col+0.5)*\s},{-(\row+0.5)*\s}) {\val};
}
\foreach \row/\col/\val in {1/3/7, 2/0/8, 2/1/9, 2/2/12, 2/3/13, 3/0/10, 3/1/11, 3/2/14}{
    \fill[green!30] ({\col*\s},{-(\row+1)*\s}) rectangle +(\s,\s);
    \draw[thin] ({\col*\s},{-(\row+1)*\s}) rectangle +(\s,\s);
    \node at ({(\col+0.5)*\s},{-(\row+0.5)*\s}) {\val};
}
\fill[blue!40] ({3*\s},{-4*\s}) rectangle +(\s,\s);
\draw[thin] ({3*\s},{-4*\s}) rectangle +(\s,\s);
\node[font=\scriptsize\bfseries] at ({3.5*\s},{-3.5*\s}) {15};
\foreach \row/\col/\val in {0/4/16, 0/5/17, 0/6/20, 0/7/21, 1/4/18, 1/5/19, 1/6/22, 1/7/23}{
    \fill[red!30] ({\col*\s},{-(\row+1)*\s}) rectangle +(\s,\s);
    \draw[thin] ({\col*\s},{-(\row+1)*\s}) rectangle +(\s,\s);
    \node at ({(\col+0.5)*\s},{-(\row+0.5)*\s}) {\val};
}
\foreach \bi in {0,1}{
  \foreach \bj in {0,...,3}{
    \draw[thick] ({\bj*2*\s},{-\bi*2*\s}) rectangle +({2*\s},{-2*\s});
  }
}
\draw[red!70!black, line width=1.5pt, dashed] ({4*\s}, 0) -- ({4*\s}, {-4*\s});
\node[font=\tiny, red!70!black, anchor=south] at ({2*\s}, 0.08) {Window 0};
\node[font=\tiny, red!70!black, anchor=south] at ({6*\s}, 0.08) {Window 1};
\node[font=\tiny, align=center, text width=2.5cm] at ({4*\s},{-4.5*\s}) {\textcolor{red!70!black}{$\boldsymbol{\times}$\, 16--23 spatially distant}};

\begin{scope}[xshift=3.5cm]
\node[font=\footnotesize\bfseries, anchor=south] at ({4*\s}, 0.4) {(b) Row-major 1D layout};

\foreach \row/\col/\val in {0/0/0, 0/1/1, 0/2/4, 0/3/5, 0/4/8, 0/5/9, 0/6/12, 0/7/13, 1/0/2, 1/1/3, 1/2/6, 1/3/7, 1/4/10, 1/5/11, 1/6/14}{
    \draw[thin] ({\col*\s},{-(\row+1)*\s}) rectangle +(\s,\s);
    \node at ({(\col+0.5)*\s},{-(\row+0.5)*\s}) {\val};
}
\fill[orange!30] ({7*\s},{-2*\s}) rectangle +(\s,\s);
\draw[thin] ({7*\s},{-2*\s}) rectangle +(\s,\s);
\node at ({7.5*\s},{-1.5*\s}) {15};
\foreach \row/\col/\val in {2/0/16, 2/1/17, 2/2/20, 2/3/21, 2/4/24, 2/5/25, 2/6/28, 2/7/29, 3/0/18, 3/1/19, 3/2/22, 3/4/26, 3/5/27, 3/6/30, 3/7/31}{
    \fill[green!30] ({\col*\s},{-(\row+1)*\s}) rectangle +(\s,\s);
    \draw[thin] ({\col*\s},{-(\row+1)*\s}) rectangle +(\s,\s);
    \node at ({(\col+0.5)*\s},{-(\row+0.5)*\s}) {\val};
}
\fill[blue!40] ({3*\s},{-4*\s}) rectangle +(\s,\s);
\draw[thin] ({3*\s},{-4*\s}) rectangle +(\s,\s);
\node[font=\scriptsize\bfseries] at ({3.5*\s},{-3.5*\s}) {23};
\foreach \bi in {0,1}{
  \foreach \bj in {0,...,3}{
    \draw[thick] ({\bj*2*\s},{-\bi*2*\s}) rectangle +({2*\s},{-2*\s});
  }
}
\node[font=\tiny, align=center, text width=2.5cm] at ({4*\s},{-4.5*\s}) {\textcolor{green!50!black}{$\boldsymbol{\checkmark}$\, 15--31 spatially coherent}};
\end{scope}

\node[font=\tiny, anchor=north] at (3.25cm,{-5.4*\s}) {%
  \tikz\fill[blue!40] (0,0) rectangle (0.18,0.18); Anchor \enspace
  \tikz\fill[green!30] (0,0) rectangle (0.18,0.18); Near \enspace
  \tikz\fill[red!30] (0,0) rectangle (0.18,0.18); Distant \enspace
  \tikz\fill[orange!30] (0,0) rectangle (0.18,0.18); Far edge};

\end{tikzpicture}
\caption{Why the 2D token ordering is incompatible with SWA. Numbers indicate position in the 1D sequence. \textbf{(a)}~Chunk-2D: \colorbox{red!30}{neighbors 16--23} map to the \textit{top} of Window~1, spatially far from the \colorbox{blue!40}{anchor}. \textbf{(b)}~Row-major 1D: nearly \colorbox{green!30}{all neighbors} fall into surrounding rows, forming a coherent neighborhood.}
\label{fig:swa2d_problem}
\end{figure}

\paragraph{Factor 2: Serialization order determines specialization quality.}
SWA increases cross-window interaction, but it still operates over a 1D token sequence.
Therefore, applying SWA to image tokens requires choosing how the 2D patch grid is serialized into that sequence.
The SWA-2D variant above keeps Qwen2.5-VL's chunk-window ordering: it groups nearby patches into local blocks, then places blocks from the same chunk next to one another in the 1D sequence.
This ordering is appropriate when attention is computed separately inside each chunk, but it becomes problematic when a sliding window moves along the sequence.
As shown in Figure~\ref{fig:swa2d_problem}a, when SWA slides across a chunk boundary under this 2D layout, neighboring sequence positions can jump to distant image regions, so a single local attention window mixes spatially near and far tokens.
We therefore test whether SWA should use a serialization that preserves spatial continuity.
To test this, we keep the SWA operator fixed and introduce \emph{SWA-1D}: a row-major 1D serialization that flattens image blocks left-to-right and top-to-bottom, making consecutive sequence positions correspond more closely to neighboring image regions (Figure~\ref{fig:arrangement}b--c).
Under this controlled change, SHS-Index increases from .588 (SWA-2D) to .600 (SWA-1D).
\textbf{This identifies serialization order as a second source of the SHS gap: for sliding-window attention on the original 2D layout, token serialization should preserve spatially coherent neighborhoods.}

\paragraph{Factor 3: Softmax distributes attention to background.}
Even after increasing cross-window interaction and improving serialization, local attention still lets each query attend to only a subset of tokens.
Within that subset, standard attention normalizes scores with softmax \citep{vaswani2017attention}, forcing the attention weights over available keys to consume the query's full probability mass.
When a local window contains many uninformative background tokens, this constraint may assign nontrivial attention to tokens that the head would otherwise prefer to ignore.
We therefore test whether local attention benefits from an explicit absorber for surplus probability mass.
Following recent efficient attention designs such as GPT-OSS \citep{openai2025gptoss}, we add a per-head learnable sink bias $s^{(h)}_{\mathrm{aux}}$: as defined in Eq.~\ref{eq:attn_weight}, it contributes an extra denominator term but has no associated image token, so it acts as a virtual sink that can absorb attention mass instead of forcing that mass onto foreground/background tokens.
The SHS evidence is mixed but informative.
Adding the sink slightly reduces SHS for SWA-1D (.600 $\to$ .598), which we report explicitly.
However, the same intervention increases SHS in the other controlled settings: chunk-2D (.577 $\to$ .579), SWA-2D (.588 $\to$ .597), and full attention (.606 $\to$ .611).
Allocation analysis further shows the sink's intended role: in SWA models, 94--97\% of heads redirect more background than foreground attention to the sink (Appendix~\ref{sec:appx_mechanism}).
\textbf{This identifies local softmax allocation as a third source of the SHS gap: local-attention softmax can waste attention on uninformative tokens, and an explicit sink provides a useful absorber channel, though its SHS effect depends on the surrounding attention structure with SWA-1D as exception.}

\begin{figure}[t]
\centering
\includegraphics[width=\columnwidth]{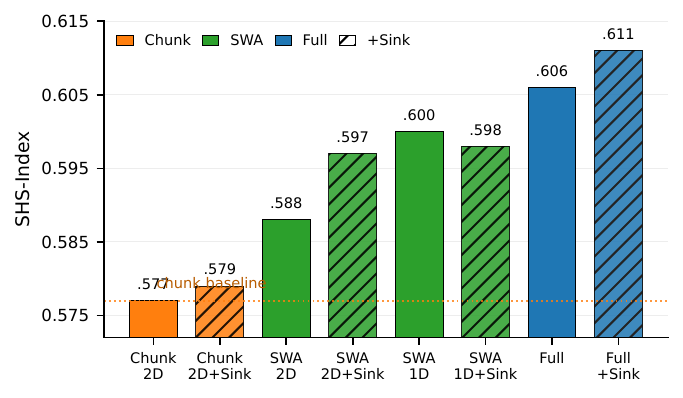}
\caption{\textbf{SHS-Index across structural configurations.} The controlled variants differ only in the local attention structure while sharing the same backbone and training recipe. SHS-Index rises under the structure changes validated above, with Full+Sink shown as the highest-SHS diagnostic point and chunk-2D as the lowest.}
\label{fig:shs_config_bar}
\end{figure}

Putting the controlled interventions together gives a compact view of how attention structure affects SHS-Index (Figure~\ref{fig:shs_config_bar}; full configurations in Table~\ref{tab:model_configs}).
Because the variants differ only in attention structure, the spread in SHS-Index directly reflects how these structural choices affect head specialization.
This makes SHS-Index a natural bridge from mechanism to design: after showing that attention structure shapes the signal, we test whether the same signal is aligned with downstream quality across the controlled configurations.

\paragraph{SHS-Index correlates with benchmark performance within the controlled design space.}
\label{sec:validation}
We evaluate the sufficiently trained diagnostic configurations on 22 downstream benchmarks, including 20 image tasks such as AI2D and OCRBench, and 2 video tasks including Video-MME.
After correcting a malformed-output issue in MMBench-EN, SHS-Index correlates with the 22-task benchmark average at 8k (Pearson $r{=}0.858$, $p{=}0.006$; exact permutation $p{=}0.008$; Figure~\ref{fig:hero_b}; full robustness analysis in Appendix~\ref{sec:appx_correlation}). The bootstrap 95\% interval is $[0.51,0.98]$.
This is a cross-architecture signal at a sufficiently trained checkpoint, not a causal or within-run predictor. It is useful for comparing attention designs in this controlled setting, while task families that depend on exact geometry or counting remain only partially captured.
Full SHS-Index trajectories are reported in Appendix~\ref{sec:appx_shs_data} (Table~\ref{tab:shs_all_iter}), and the complete benchmark trajectories used in the architectural comparisons are reported in Appendix~\ref{sec:appx_bench} (Tables~\ref{tab:bench_iter} and~\ref{tab:bench_iter_22} for 20-image-task and 22-task averages, respectively). Collective-head evidence is in Appendix~\ref{sec:appx_collective}.

\section{SHS-Guided Design of Ariadne Attention}
\label{sec:design}

The previous section identifies structural factors that shape SHS and shows that SHS-Index correlates with downstream benchmark performance in the controlled comparison.
We now use these findings to design a hybrid attention pattern.
Guided by the validated components---sliding-window attention, 1D serialization, and sink bias---we further increase token interaction across image dimensions by alternating row- and column-major serialization.
The resulting hybrid, \textbf{Ariadne Attention}, is a validation of the SHS framework rather than a claim of a new primitive attention operator.

\begin{figure}[!t]
\centering
\includegraphics[width=\columnwidth]{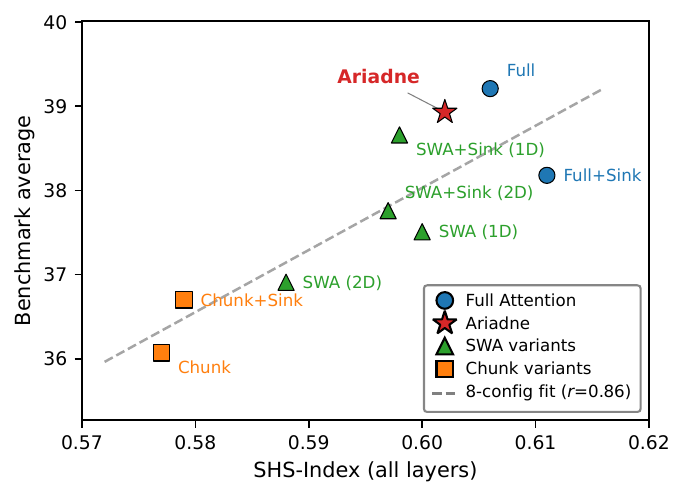}
\caption{\textbf{SHS-Index and benchmark quality.} Across the eight diagnostic configurations, SHS-Index correlates with the corrected 22-task average ($r{=}0.858$, $p{=}0.006$; dashed line = 8-config linear fit). Ariadne is shown as a later design point and is not used to define the primary fit.}
\label{fig:hero_b}
\end{figure}

\subsection{Ariadne Attention}
\label{sec:exp_overview}
\label{sec:ariadne}

We now instantiate the design suggested by the SHS analysis.
Ariadne starts from the three components that improved or stabilized SHS in the controlled study: sliding-window attention (SWA) to relax chunk isolation, spatially coherent 1D serialization to make local windows meaningful on images, and a per-head sink bias to absorb surplus softmax mass.
SWA replaces non-overlapping chunk windows with overlapping token-wise neighborhoods, so adjacent windows can exchange information without returning to full attention.
The sink bias uses FlashAttention-3's native \texttt{s\_aux} term \citep{shah2024flashattention3} as a zero-token absorber, serving the same role as attention sinks/register tokens \citep{xiao2024streamingllm,darcet2024vision} without adding extra visual tokens.

\begin{figure}[t]
\centering
\definecolor{rowc}{RGB}{99,166,226}%
\definecolor{colc}{RGB}{248,177,73}%
\definecolor{fullc}{RGB}{214,96,96}%
\begin{tikzpicture}[
  font=\scriptsize,
  >=stealth,
  layer/.style={minimum width=3.8cm, minimum height=0.38cm, rounded corners=2pt,
                inner sep=0pt, font=\tiny\bfseries, text=white, draw=none},
  row layer/.style={layer, fill=rowc},
  col layer/.style={layer, fill=colc},
  full layer/.style={layer, fill=fullc},
  arrow/.style={->, thick, gray!70},
  block brace/.style={decorate, decoration={brace, mirror, amplitude=5pt}, thick, gray!60},
]

\def\lh{0.50}

\node[font=\footnotesize\bfseries, anchor=south] at (0, 8*\lh + 0.2) {One Block $\times 4$ = 32 Layers};

\node[full layer] (F) at (0, 7*\lh) {Full Attention};
\node[col layer] (C6) at (0, 6*\lh) {Col-major SWA + Sink};
\node[col layer] (C5) at (0, 5*\lh) {Col-major SWA + Sink};
\node[col layer] (C4) at (0, 4*\lh) {Col-major SWA + Sink};
\node[row layer] (R3) at (0, 3*\lh) {Row-major SWA + Sink};
\node[row layer] (R2) at (0, 2*\lh) {Row-major SWA + Sink};
\node[row layer] (R1) at (0, 1*\lh) {Row-major SWA + Sink};
\node[row layer] (R0) at (0, 0)     {Row-major SWA + Sink};

\foreach \i [evaluate=\i as \j using int(\i+1)] in {0,...,6}{
  \pgfmathsetmacro{\ya}{\i*\lh + 0.19}
  \pgfmathsetmacro{\yb}{\j*\lh - 0.19}
  \draw[arrow] (0, \ya) -- (0, \yb);
}

\draw[block brace] (-2.1, 3*\lh+0.19) -- (-2.1, -0.19)
  node[midway, left=6pt, font=\tiny, text=gray, align=right] {4 layers\\row-major};
\draw[block brace] (-2.1, 6*\lh+0.19) -- (-2.1, 4*\lh-0.19)
  node[midway, left=6pt, font=\tiny, text=gray, align=right] {3 layers\\col-major};
\draw[block brace] (-2.1, 7*\lh+0.19) -- (-2.1, 7*\lh-0.19)
  node[midway, left=6pt, font=\tiny, text=gray] {1 full};

\draw[arrow] (0, -0.55) -- (0, -0.19) node[midway, right=2pt, font=\tiny, gray] {input};
\draw[arrow] (0, 7*\lh+0.19) -- (0, 7*\lh+0.55) node[midway, right=2pt, font=\tiny, gray] {output};

\end{tikzpicture}
\caption{The \textbf{4-row-3-col} block architecture of Ariadne Attention. Each block contains 4 row-major SWA~+~sink layers, 3 column-major SWA~+~sink layers, and 1 full-attention layer; four repeats form the 32-layer ViT.}
\label{fig:4r3c}
\end{figure}
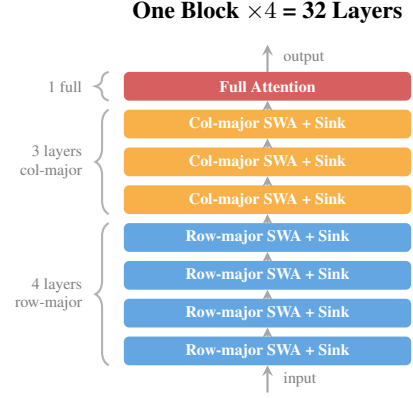

The remaining design question is how to serialize 2D image tokens for SWA.
A single row-major order preserves horizontal continuity but propagates information less directly across columns.
Ariadne therefore alternates row-major and column-major serialization across depth, so local windows connect both image dimensions over successive layers.
Concretely, each 8-layer block contains 4 row-major SWA~+~sink layers, 3 column-major SWA~+~sink layers, and 1 full-attention anchor layer; repeating this block four times forms the 32-layer ViT (Figure~\ref{fig:4r3c}).
The 4:3 ratio is a practical refinement rather than a headline contribution: an 8-layer block already propagates information along both image axes, so nearby ratios should have similar receptive-field coverage. We use block-wise RRRR-CCC ordering because per-layer RCRC alternation would require a memory-bound token re-permutation at every layer, whereas RRRR-CCC re-permutes only at row--column boundaries.
We set the SWA window to $w{=}64$ per side (128 tokens total context), matching GPT-OSS \citep{openai2025gptoss}; an ablation over $w{\in}\{32,64,128\}$ selects the same value as best (SHS$=$.602, 22-task score $=$38.93) (Appendix~\ref{sec:appx_models}).

\subsection{Main Results}
\label{sec:results}

\begin{figure}[t]
\centering
\includegraphics[width=\columnwidth]{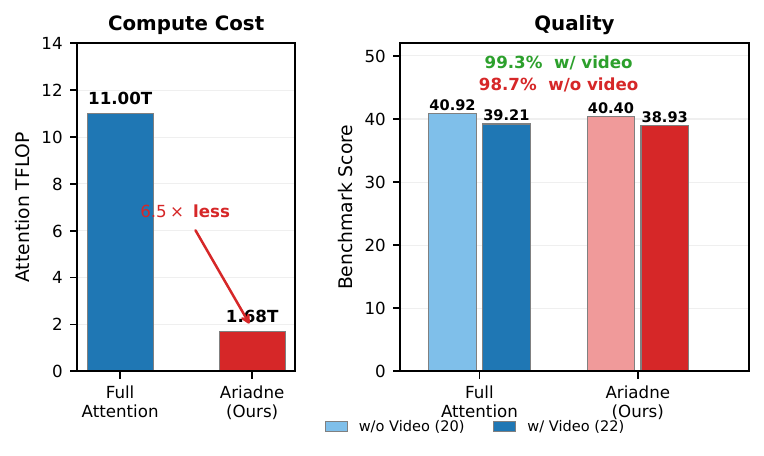}
\caption{Attention compute vs.\ image-task quality at $896^2$. Ariadne (\textcolor{red}{$\bigstar$}), an SHS-informed hybrid, comes within about 0.5 points of full attention at $6.5\times$ lower attention FLOPs.}
\label{fig:tflops}
\end{figure}

\begin{table}[t]
\centering
\small
\setlength{\tabcolsep}{5pt}
\resizebox{\columnwidth}{!}{%
\begin{tabular}{lcc}
\toprule
& \textbf{Image tasks (20)} & \textbf{Video tasks (2)} \\
\midrule
Full Attention & \textbf{40.92} & 22.1 \\
Chunk-window Attention & 37.35 & 23.3 \\
Ariadne Attention & 40.40 & \textbf{24.3} \\
\bottomrule
\end{tabular}
}
\caption{Corrected benchmark averages reported separately for image and video tasks at the 8k primary checkpoint. Ariadne is within 0.52 points of full attention on image tasks; the two video tasks are reported separately rather than folded into the headline claim.}
\label{tab:hero_bench}
\end{table}

We first ask whether the SHS-guided design closes the quality gap to full attention.
Ariadne scores 40.40 on the 20-task image benchmark versus 40.92 for full attention, a 0.52-point gap, and reaches 24.3 on the two video tasks (Table~\ref{tab:hero_bench}; Figure~\ref{fig:tflops}).
Relative to chunk-window attention, Ariadne improves the corrected 20-image average by $+3.05$, showing that the hybrid recovers most of full attention's aggregate image quality while improving over the chunk-window baseline.

The largest gains occur on tasks where chunk-window isolation is most limiting: tasks that require finding evidence across distant regions or combining fine local details with broader context.
On perception benchmarks, Ariadne improves over chunk-window attention by $+6.1$ on CV-Bench and $+7.2$ on VLMs-are-Blind.
On high-resolution visual search, V* improves by $+11.0$, consistent with row/column serialization and sliding windows improving evidence propagation across the image.
OCR and document-style tasks show the same pattern: DocVQA improves by $+9.8$, InfoVQA by $+7.7$, and OCRBench by $+8.9$, reflecting the need to connect local text evidence with page-level layout context across distant regions.

The remaining task-level differences are also informative.
Grounding tasks change little (RefCOCO $+0.2$, RefCOCO+ $\pm0.0$, RefCOCOg $+1.1$), suggesting that these benchmarks are less sensitive to the cross-region integration targeted by Ariadne and are dominated by local phrase--region matching.
Chart and counting tasks regress slightly (ChartXiv-RQ $-1.6$, ChartQA $-1.2$, PixMo-Count $-1.9$), indicating that exact geometric reasoning and instance-level counting may require additional structure-preserving mechanisms beyond the current row/column serialization.
Overall, Ariadne's strengths are most pronounced on tasks that require cross-region evidence integration, while highly geometry- or count-sensitive tasks point to a complementary direction for future hybrid designs.
Full per-task results across all 22 tasks are reported in Appendix~\ref{sec:appx_bench} (Table~\ref{tab:per_task}).

\subsection{Efficiency}
\label{sec:efficiency}

\begin{figure}[t]
\centering
\includegraphics[width=\columnwidth]{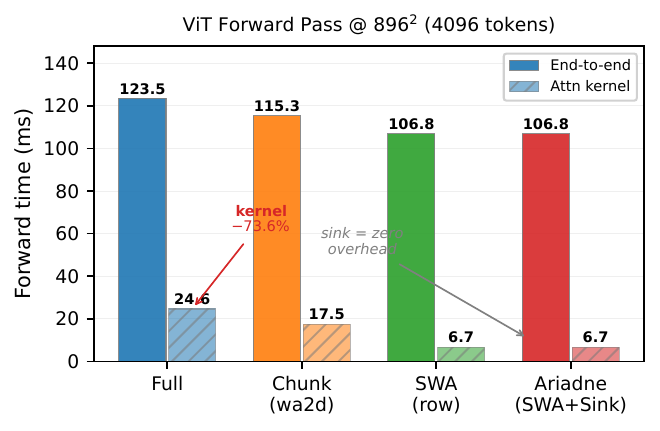}
\caption{ViT forward-pass time at $896^2$ (4{,}096 tokens). Solid bars: end-to-end block time; hatched bars: attention kernel only. In the updated scaling run, Ariadne is 106.8\,ms end-to-end versus 123.5\,ms for Full, and its attention kernel is 6.7\,ms versus 24.6\,ms.}
\label{fig:efficiency_bar}
\end{figure}

Finally, we test actual kernel timings, not just asymptotic attention complexity.
Sink bias adds zero computational overhead: FlashAttention-3 implements \texttt{s\_aux} as a single scalar addition to the softmax denominator \citep{shah2024flashattention3}.
At $896^2$, Ariadne's end-to-end ViT forward time is 106.8\,ms vs.\ 123.5\,ms for full attention ($-13.5\%$); at $1792^2$, the saving grows to $-39.4\%$ as the quadratic cost becomes dominant.
At $896^2$, SWA is also faster than chunk-window attention (6.7\,ms vs.\ 17.5\,ms for the attention kernel) because FlashAttention-3's contiguous SWA kernel is better optimized than the varlen interface (Figure~\ref{fig:efficiency_bar}).
Thus Ariadne improves the measured quality--compute frontier; Appendix~\ref{sec:appx_efficiency} reports timing across seven resolutions and distinguishes attention FLOPs from end-to-end block time.

\section{Related Work}
\label{sec:related}

\paragraph{Attention head behavior.}
Prior work identifies specialized attention heads for syntactic and positional roles in NLP \citep{voita2019analyzing,michel2019sixteen}. In vision, self-supervised ViTs develop object-segmenting CLS-attention maps \citep{caron2021emerging}, while CLIP and generative multimodal transformers show semantically interpretable head roles \citep{gandelsman2023clip,basile2025headpursuit}. We extend this line to supervised multimodal ViTs with SHS-Index, a continuous, direction-agnostic foreground/background measure. Our result is a cross-architecture correlation at a fixed, sufficiently trained checkpoint, not a claim that specialization monotonically predicts every task.

\paragraph{Hybrid attention in Vision Transformers.}
Efficient ViTs span shifted windows \citep{liu2021swin}, axial/stripe designs \citep{ho2019axial,wang2020axialdeeplab,huang2019ccnet,dong2021cswin}, and local/hybrid backbones such as Twins, Focal, NAT, and MaxViT \citep{chu2021twins,yang2021focal,hassani2022nat,tu2022maxvit}. Vision state-space models further show that 2D-to-1D scan order matters \citep{liu2024vmamba,zhu2024vim}. InternVL~2.5 and Qwen3-VL use full attention, whereas Qwen2.5-VL uses chunk windows \citep{chen2024internvl25,bai2025qwen3vl,bai2025qwen25vl}. We study this trade-off inside a controlled multimodal-LLM pipeline. Ariadne is functionally related to axial/stripe attention, but uses overlapping windows and a block-wise row/column schedule for fused-kernel efficiency; the central contribution is SHS, not a new primitive. Its sink bias also relates to ViT registers and LLM attention sinks \citep{darcet2024vision,xiao2024streamingllm}.

\section{Conclusion}
\label{sec:conclusion}

This paper studies why hybrid ViT attention lags behind full attention in multimodal LLMs. We discover Semantic Head Specialization (SHS): full-attention heads separate into object- and background-specialist roles, while chunk-window heads show weaker separation. SHS-Index quantifies this behavior, separates full and chunk families across 16 open-source visual encoders and VLMs, and exposes three structural factors: window interaction, token serialization, and local softmax allocation. At 8k, it correlates with the corrected 22-task average ($r=.858$), with an intermediate rather than monotonic optimum. Guided by these factors, Ariadne reaches 40.40 versus 40.92 for full attention on 20 image tasks, at $6.5\times$ lower attention FLOPs and 13.5\% lower end-to-end ViT time at $896^2$. SHS-Index is therefore a compact diagnostic for comparing and designing hybrid ViT attention within this scope.

\section*{Limitations}

\paragraph{Scale of the controlled study.}
All nine architectures share the same training configuration (random-init ViT, Qwen2-0.5B, 10k steps), use one random seed, and are evaluated at a single primary checkpoint.
The corrected MMBench-EN evaluation is available for the 8k primary checkpoint; the 10k sensitivity rows retain the original evaluation and are labeled accordingly. The controlled design isolates attention operators, but the quantitative correlation has not been tested with a larger language backbone, multiple seeds, or from-scratch comparisons to every axial/stripe baseline. The open-source analysis now spans roughly five independent full-attention ViT lineages and supports the structural separation at scale, but it does not establish benchmark correlation across those heterogeneous models.

\section*{Acknowledgments}

This work was supported by the National Key Research and Development Program of China under Grant No. 2024YFB2906603, and in part by the National Natural Science Foundation of China (NSFC) (No. 62372009).

\bibliography{references}

\appendix

\section{Model Architecture and Configurations}
\label{sec:appx_models}

\subsection{ViT Architecture}
\label{sec:appx_vit_arch}

All nine models use a 32-layer ViT encoder following the Qwen2.5-VL \citep{bai2025qwen25vl} design.
Of the 32 layers, 4 (at 0-indexed positions 7, 15, 23, 31) use full self-attention; the remaining 28 use the local attention operator under study.
The ViT extracts non-overlapping $14 \times 14$ patches and groups every $2 \times 2$ spatially adjacent patches into an atomic block of 4 sub-tokens, the basic unit of both chunk-window grouping and 1D serialization.
Architectural hyperparameters are listed in Table~\ref{tab:vit_arch}.

\begin{table}[ht!]
\centering
\small
\setlength{\tabcolsep}{4pt}
\begin{tabular}{@{}ll@{}}
\toprule
\textbf{Parameter} & \textbf{Value} \\
\midrule
Layers & 32 \\
Full-attention layers & positions 7, 15, 23, 31 \\
Local-attention layers & 28 \\
Hidden dimension & 1{,}280 \\
MLP dimension & 3{,}456 \\
Attention heads & 16 \\
Head dimension & 80 \\
Patch size & $14 \times 14$ \\
Block size & $2 \times 2$ patches (4 sub-tokens) \\
Position encoding & 2D RoPE \\
Initialization & Random \\
\bottomrule
\end{tabular}
\caption{ViT encoder hyperparameters, shared across all 9 configurations.}
\label{tab:vit_arch}
\end{table}

\subsection{Nine Model Configurations}

Table~\ref{tab:model_configs} lists all nine attention configurations explored in this work. They span three design axes: token layout (2D vs.\ 1D), attention type (chunk vs.\ sliding window vs.\ full), and sink bias (present or absent).

\begin{table*}[t]
\centering
\small
\begin{tabular}{lcccccc}
\toprule
\textbf{Short Name} & \textbf{Layout} & \textbf{Attention} & \textbf{Sink} & \textbf{Window Size} & \textbf{SHS-Index} & \textbf{Performance (20-image)} \\
\midrule
wa2d & 2D & Chunk & No & 64 tok & .577 & 37.35 \\
wa2d\_sink & 2D & Chunk & Yes & 64 tok & .579 & 37.94 \\
swa2d & 2D & SWA & No & 128 tok & .588 & 38.09 \\
swa2d\_sink & 2D & SWA & Yes & 128 tok & .597 & 39.13 \\
swa\_row & 1D row & SWA & No & 128 tok & .600 & 38.80 \\
swa\_sink\_row & 1D row & SWA & Yes & 128 tok & .598 & 40.14 \\
swa\_sink\_4r3c & 1D row+col & SWA & Yes & 128 tok & .602 & 40.40 \\
full & N/A & Full & No & Global & .606 & 40.92 \\
full\_sink & N/A & Full & Yes & Global & .611 & 39.95 \\
\bottomrule
\end{tabular}
\caption{Complete model configurations. All models share the same ViT architecture (32 layers, 16 heads, $d_h{=}80$) and training setup (Table~\ref{tab:training_config}); only the local attention operator differs. Bench = macro-average over 20 image tasks.}
\label{tab:model_configs}
\end{table*}

\subsection{Attention Design Space}

\paragraph{Chunk window attention.}
Blocks are grouped into $4 \times 4$-block windows ($W{=}64$ tokens per window), similar to the shifted-window mechanism of Swin Transformer \citep{liu2021swin}. Self-attention is computed independently within each window. Tokens in different windows can only exchange information through the 4 full-attention layers.

\paragraph{Sliding window attention (SWA).}
Each token attends to a window of $(w_L, w_R)$ tokens to its left and right in the 1D sequence, as popularized by Mistral \citep{jiang2023mistral}. We use $w_L{=}w_R{=}64$ (128 tokens total context). Unlike chunk windows, SWA windows overlap, enabling cross-window information flow. Implemented via FlashAttention-3 \citep{shah2024flashattention3} varlen interface.

\paragraph{Sink bias.}
A single learnable scalar $s_{\text{aux}}^{(h)}$ per head is added to the softmax denominator, acting as a virtual sink token that absorbs surplus attention mass \citep{xiao2024streamingllm, darcet2024vision}. Natively supported by FlashAttention-3's \texttt{s\_aux} parameter at zero additional cost.

\paragraph{Token serialization.}
For 1D-layout models, blocks are serialized into a flat sequence. Row-major: left-to-right, top-to-bottom. Column-major: top-to-bottom, left-to-right. The Ariadne configuration alternates 4 row-major and 3 column-major layers within each block of 7 local layers (4-row-3-col pattern). See Figure~\ref{fig:arrangement} for an illustration.

\subsection{Window Size Ablation}

Table~\ref{tab:window_ablation} compares three window sizes on the Ariadne configuration.

\begin{table}[ht!]
\centering
\small
\begin{tabular}{lcccc}
\toprule
\textbf{Window} & \textbf{Context} & \textbf{SHS} & \textbf{20-image} & \textbf{22-task} \\
\midrule
$w{=}32$ & 64 tok & .597 & 38.11 & 36.95 \\
$w{=}64$ & 128 tok & \textbf{.602} & \textbf{40.40} & \textbf{38.93} \\
$w{=}128$ & 256 tok & .595 & 38.65 & 37.27 \\
\bottomrule
\end{tabular}
\caption{Window size ablation (Ariadne, swa\_sink\_4r3c) at the primary checkpoint. $w{=}64$ is optimal under both 20-image and 22-task averages, matching the window size used by GPT-OSS \citep{openai2025gptoss} and MiMo-V2-Flash \citep{mimo2026v2flash}.}
\label{tab:window_ablation}
\end{table}

\section{Complete Benchmark Results}
\label{sec:appx_bench}

This appendix reports full benchmark numbers across all iterations (Tables~\ref{tab:bench_iter} and~\ref{tab:bench_iter_22}, for the 20-image-task and 22-task averages, respectively) and per-task breakdowns for the three key models at the primary checkpoint (Table~\ref{tab:per_task}).

\subsection{Benchmark Across Iterations}

\begin{figure}[t]
\centering
\includegraphics[width=\columnwidth]{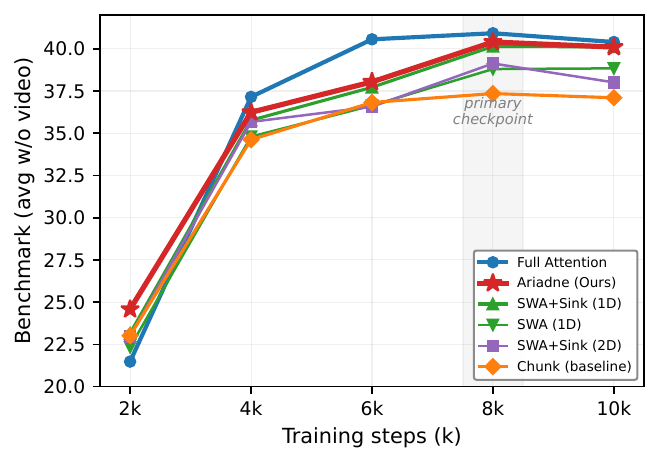}
\caption{Benchmark trajectory across training for six key models. Architectural rankings are stable across late checkpoints, while SWA+Sink (2D) begins to overfit in the late stage.}
\label{fig:bench_iter}
\end{figure}

Table~\ref{tab:bench_iter} reports the macro-average benchmark (20 image tasks, excluding video) at each checkpoint; Table~\ref{tab:bench_iter_22} reports the 22-task counterpart used in the SHS-quality correlation (\S\ref{sec:validation}). The ranking remains stable across late checkpoints under both metrics, motivating our choice of primary checkpoint; see Figure~\ref{fig:bench_iter} for a visualization.

\subsection{Key Models Per-Task Results}

Table~\ref{tab:per_task} breaks down the benchmark into all 22 individual tasks for the three key models (Full, Ariadne, Chunk). Ariadne recovers most of the gap on vision-centric tasks while grounding tasks show near-parity.


\begin{table*}[t]
\centering
\small
\begin{tabular}{lccccccccc}
\toprule
\textbf{Iter} & \textbf{full\_s$^\dagger$} & \textbf{full} & \textbf{4r3c} & \textbf{sink\_r} & \textbf{swa\_r} & \textbf{s2d\_s} & \textbf{swa2d} & \textbf{w\_s} & \textbf{wa2d} \\
\midrule
2k & 22.25 & 21.48 & 24.58 & 23.15 & 22.34 & 22.98 & 22.67 & 23.22 & 23.02 \\
4k & 34.44 & 37.15 & 36.24 & 35.77 & 34.80 & 35.67 & 33.30 & 35.48 & 34.62 \\
6k & 37.70 & 40.56 & 38.04 & 37.72 & 36.64 & 36.57 & 37.64 & 38.19 & 36.82 \\
8k & 39.95 & 40.92 & 40.40 & 40.14 & 38.80 & 39.13 & 38.09 & 37.94 & 37.35 \\
10k & 39.94 & 40.40 & 40.10 & 40.11 & 38.84 & 38.01 & 35.21 & 38.27 & 37.10 \\
\bottomrule
\end{tabular}
\caption{Benchmark (avg w/o video, 20 tasks) across training iterations. The 8k primary checkpoint uses corrected MMBench-EN scores; the 10k row retains the original evaluation because the re-evaluation was run at 8k only. $^\dagger$full\_sink 2k--6k based on general tasks only (10 tasks).}
\label{tab:bench_iter}
\end{table*}

\begin{table*}[t]
\centering
\small
\begin{tabular}{lccccccccc}
\toprule
\textbf{Iter} & \textbf{full\_s$^\dagger$} & \textbf{full} & \textbf{4r3c} & \textbf{sink\_r} & \textbf{swa\_r} & \textbf{s2d\_s} & \textbf{swa2d} & \textbf{w\_s} & \textbf{wa2d} \\
\midrule
2k & 22.25 & 21.42 & 24.41 & 23.04 & 22.35 & 22.85 & 22.58 & 23.13 & 22.92 \\
4k & 33.25 & 35.77 & 35.16 & 34.67 & 33.73 & 34.67 & 32.41 & 34.49 & 33.61 \\
6k & 36.26 & 38.93 & 36.85 & 36.53 & 35.55 & 35.45 & 36.54 & 37.01 & 35.62 \\
8k & 38.18 & 39.21 & 38.93 & 38.66 & 37.51 & 37.76 & 36.91 & 36.70 & 36.07 \\
10k & 38.20 & 38.73 & 38.66 & 38.63 & 37.61 & 36.84 & 34.27 & 37.02 & 35.91 \\
\bottomrule
\end{tabular}
\caption{Benchmark (avg w/ video, 22 tasks) across training iterations. Includes Video-MME and Video-MMMU. The 8k primary checkpoint uses corrected MMBench-EN scores; the 10k row retains the original evaluation. $^\dagger$full\_sink 2k--6k based on general tasks only (10 tasks).}
\label{tab:bench_iter_22}
\end{table*}

\begin{table*}[t]
\centering
\small
\setlength{\tabcolsep}{5pt}
\begin{tabular}{@{}llrrrr@{}}
\toprule
\textbf{Cat.} & \textbf{Task} & \textbf{Full} & \textbf{Chunk} & \textbf{Ariadne} & $\boldsymbol{\Delta}$ \\
\midrule
\multirow{10}{*}{General}
& AI2D           & 37.2 & 35.3 & \textbf{37.5} & $+2.3$ \\
& BLINK          & 37.0 & \textbf{39.4} & 38.3 & $-1.1$ \\
& CV-Bench       & 47.7 & 43.4 & \textbf{49.5} & $+6.1$ \\
& MMMU           & 29.1 & 28.7 & \textbf{30.2} & $+1.5$ \\
& MMBench-EN     & 36.34 & 36.94 & \textbf{38.57} & $+1.63$ \\
& MMBench-CN     & 30.8 & 27.2 & \textbf{29.9} & $+2.7$ \\
& Mantis         & 16.1 & 15.2 & \textbf{22.1} & $+6.9$ \\
& MME-RW-CN      & 26.8 & 27.2 & \textbf{27.5} & $+0.3$ \\
& V*             & 48.7 & 28.8 & \textbf{39.8} & $+11.0$ \\
& VLMs-are-Blind & 31.5 & 28.2 & \textbf{35.4} & $+7.2$ \\
\midrule
\multirow{6}{*}{OCR}
& ChartXiv-RQ    & 19.1 & \textbf{21.0} & 19.4 & $-1.6$ \\
& ChartXiv-DQ    & 52.1 & \textbf{50.2} & 49.7 & $-0.5$ \\
& DocVQA         & 53.5 & 39.8 & \textbf{49.5} & $+9.8$ \\
& InfoVQA        & 32.4 & 23.6 & \textbf{31.3} & $+7.7$ \\
& ChartQA        & 19.4 & \textbf{18.8} & 17.7 & $-1.2$ \\
& OCRBench       & 68.0 & 54.7 & \textbf{63.6} & $+8.9$ \\
\midrule
\multirow{3}{*}{Ground.}
& RefCOCO        & 76.2 & 75.0 & \textbf{75.2} & $+0.2$ \\
& RefCOCO+       & 63.4 & \textbf{62.8} & 62.8 & $\pm0.0$ \\
& RefCOCOg       & 67.7 & 68.5 & \textbf{69.5} & $+1.1$ \\
\midrule
Count & PixMo-Count & 25.1 & \textbf{22.3} & 20.4 & $-1.9$ \\
\midrule
\multirow{2}{*}{Video}
& Video-MME      & 27.5 & 30.9 & \textbf{32.3} & $+1.4$ \\
& Video-MMMU     & 16.7 & 15.8 & \textbf{16.2} & $+0.4$ \\
\midrule
\multicolumn{2}{@{}l}{\textbf{Avg}} & \textbf{39.21} & 36.07 & \textbf{38.93} & $+2.86$ \\
\bottomrule
\end{tabular}
\caption{Per-task benchmark results for the three key models at the 8k primary checkpoint. MMBench-EN was re-evaluated with corrected output parsing. $\Delta$ = Ariadne $-$ Chunk. Bold = best between Chunk and Ariadne. Ariadne recovers most of the full-hybrid gap on vision-centric tasks (V* $+11.0$, DocVQA $+9.8$, OCRBench $+8.9$); grounding tasks, which rely on local spatial reasoning within the sliding window, show near-parity.}
\label{tab:per_task}
\end{table*}


\section{Experimental Setup Details}
\label{sec:appx_setup}

\subsection{Training Configuration}

Table~\ref{tab:training_config} summarizes the training setup shared across all nine models; only the attention operator differs.

\begin{table}[ht!]
\centering
\small
\setlength{\tabcolsep}{4pt}
\begin{tabular}{@{}ll@{}}
\toprule
\textbf{Parameter} & \textbf{Value} \\
\midrule
ViT encoder & Random-init, 32 layers \\
LLM backbone & Qwen2-0.5B \citep{yang2024qwen2} \\
Projector & Random-init \\
ViT / LLM lr & $5{\times}10^{-4}$ / $1{\times}10^{-5}$ \\
LR schedule & Constant w/ 1k warmup \\
Global batch size & 2{,}048 \\
Max seq length & 32{,}768 tokens \\
Seq packing & Yes \citep{chen2024internvl25} \\
Steps & 10{,}000 \\
Precision & bfloat16 \\
Compute budget & $\sim$100k GPU-hours (NVIDIA Hopper) \\
Framework & Megatron-based (in-house) \\
Inference & SGLang 0.4 (modified) \\
\bottomrule
\end{tabular}
\caption{Training configuration shared across all nine models.}
\label{tab:training_config}
\end{table}

\paragraph{Training data.}
We train on \textbf{Cambrian-10M} \citep{tong2024cambrian}, a publicly available multimodal instruction-tuning dataset of approximately 10M examples spanning OCR, general vision-language understanding, math, code, science, and multi-turn conversation. We use the released category-balanced mixture without modification.

\paragraph{Design rationale.}
We deliberately choose a small LLM backend. As noted by \citet{tong2024cambrian}, a strong language model can act as a shortcut that compensates for deficiencies in visual representations, masking real differences between vision encoder designs. A weaker LLM forces the ViT to develop genuinely better representations, making architectural differences more visible. This rationale is further supported by the scaling consistency findings of \citet{zohar2024apollo}.

\subsection{Benchmark Suite}

We evaluate on 22 tasks spanning 5 capability dimensions (Table~\ref{tab:bench_suite}). The primary metric is the macro-average over the 20 image tasks (``Avg w/o video''); we additionally report the 22-task average including video (``Avg w/ video'').

\begin{table}[ht!]
\centering
\small
\setlength{\tabcolsep}{4pt}
\begin{tabular}{@{}lp{0.62\columnwidth}c@{}}
\toprule
\textbf{Cat.} & \textbf{Tasks} & \textbf{N} \\
\midrule
General & AI2D, BLINK, CV-Bench, MMMU, MMBench-EN/CN, Mantis, MME-RealWorld-CN, V*, VLMs-are-Blind & 10 \\
\midrule
OCR & ChartXiv-RQ/DQ, DocVQA, InfoVQA, ChartQA, OCRBench & 6 \\
\midrule
Ground. & RefCOCO, RefCOCO+, RefCOCOg & 3 \\
\midrule
Count & PixMo-Count & 1 \\
\midrule
Video & Video-MME, Video-MMMU & 2 \\
\bottomrule
\end{tabular}
\caption{Benchmark suite: 20 image + 2 video tasks. Video tasks excluded from primary ``Avg w/o video''; included in ``Avg w/ video.'' Citations: General \citep{kembhavi2016ai2d, fu2024blink, tong2024cambrian, yue2023mmmu, liu2023mmbench, jiang2024mantis, zhang2024mmerealworld, wu2023vstar, rahmanzadehgervi2024vlmsblind}; OCR \citep{wang2024charxiv, mathew2020docvqa, mathew2021infographicvqa, masry2022chartqa, liu2023ocrbench}; Grounding \citep{yu2016refcoco}; Count \citep{deitke2024molmo}; Video \citep{fu2024videomme, hu2025videommmu}.}
\label{tab:bench_suite}
\end{table}

\subsection{Analysis Data}

Per-head attention analysis uses images sampled from COCO val2017, paired with a foreground/background segmentation that distinguishes salient objects from the background. Each ViT token is labeled foreground or background by the dominant region within its image patch; patches crossing the boundary (edge patches) are excluded from analysis.

\section{Additional Attention Heatmaps}
\label{sec:appx_heatmaps}

This appendix provides additional received-attention heatmaps beyond the dog image in Figure~\ref{fig:hero}.
Figure~\ref{fig:dog_boxplot} shows per-head box plots for the dog image, and Figure~\ref{fig:appx_heatmaps} presents a compact 2$\times$2 grid of four additional images (bird, two\_cats, coco\_537964, coco\_043816), each comparing Full Attention, Ariadne, and Chunk Window at layers L11 and L17.

\newcommand{\heatmapfig}[3]{%
\begin{figure*}[ht]
\centering
\setlength{\tabcolsep}{1.5pt}
\renewcommand{\arraystretch}{0.85}
\begin{tabular}{@{}c ccc c@{}}
\small\textbf{Input} & \small\textbf{Full Attention} & \small\textbf{Ariadne} & \small\textbf{Chunk Window} & \\[2pt]
\includegraphics[width=0.215\textwidth]{figures/hero/cells/#1_original.jpg} &
\includegraphics[width=0.215\textwidth]{figures/hero/cells/#1_L11_full.jpg} &
\includegraphics[width=0.215\textwidth]{figures/hero/cells/#1_L11_swa.jpg} &
\includegraphics[width=0.215\textwidth]{figures/hero/cells/#1_L11_wa2d.jpg} &
\rotatebox{90}{\small\;\;L11} \\[1.5pt]
\includegraphics[width=0.215\textwidth]{figures/hero/cells/#1_seg.jpg} &
\includegraphics[width=0.215\textwidth]{figures/hero/cells/#1_L17_full.jpg} &
\includegraphics[width=0.215\textwidth]{figures/hero/cells/#1_L17_swa.jpg} &
\includegraphics[width=0.215\textwidth]{figures/hero/cells/#1_L17_wa2d.jpg} &
\rotatebox{90}{\small\;\;L17} \\
\end{tabular}
\caption{#3}
\label{#2}
\end{figure*}
}

\newcommand{\boxplotfig}[3]{%
\begin{figure*}[ht]
\centering
\setlength{\tabcolsep}{1.5pt}
\renewcommand{\arraystretch}{0.85}
\begin{tabular}{@{}ccc c@{}}
\small\textbf{Full Attention} & \small\textbf{Ariadne} & \small\textbf{Chunk Window} & \\[2pt]
\includegraphics[width=0.31\textwidth]{figures/hero/cells/#1_boxplot_L11_full.jpg} &
\includegraphics[width=0.31\textwidth]{figures/hero/cells/#1_boxplot_L11_swa.jpg} &
\includegraphics[width=0.31\textwidth]{figures/hero/cells/#1_boxplot_L11_wa2d.jpg} &
\rotatebox{90}{\small\;\;L11} \\[1.5pt]
\includegraphics[width=0.31\textwidth]{figures/hero/cells/#1_boxplot_L17_full.jpg} &
\includegraphics[width=0.31\textwidth]{figures/hero/cells/#1_boxplot_L17_swa.jpg} &
\includegraphics[width=0.31\textwidth]{figures/hero/cells/#1_boxplot_L17_wa2d.jpg} &
\rotatebox{90}{\small\;\;L17} \\
\end{tabular}
\caption{#3}
\label{#2}
\end{figure*}
}

\boxplotfig{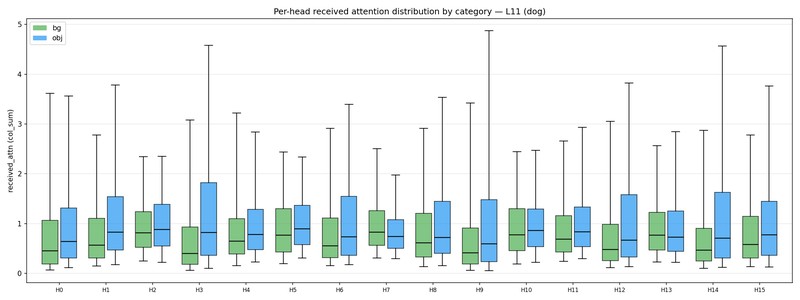}{fig:dog_boxplot}{%
\textbf{Per-head received-attention box plots} corresponding to Figure~\ref{fig:hero}. Each subplot shows the distribution for all 16 heads, split by foreground (blue) and background (green). Full Attention and Ariadne show clear obj/bg separation; Chunk Window distributions are uniform, indicating suppressed specialization.}

\newcommand{\heatmapblock}[1]{%
\setlength{\tabcolsep}{1pt}%
\renewcommand{\arraystretch}{0.85}%
\begin{tabular}{@{}cccc c@{}}
\scriptsize\textbf{Input} & \scriptsize\textbf{Full} & \scriptsize\textbf{Ariadne} & \scriptsize\textbf{Chunk} & \\[1pt]
\includegraphics[width=0.108\textwidth]{figures/hero/cells/#1_original.jpg} &
\includegraphics[width=0.108\textwidth]{figures/hero/cells/#1_L11_full.jpg} &
\includegraphics[width=0.108\textwidth]{figures/hero/cells/#1_L11_swa.jpg} &
\includegraphics[width=0.108\textwidth]{figures/hero/cells/#1_L11_wa2d.jpg} &
\rotatebox{90}{\scriptsize\;L11} \\[1pt]
\includegraphics[width=0.108\textwidth]{figures/hero/cells/#1_seg.jpg} &
\includegraphics[width=0.108\textwidth]{figures/hero/cells/#1_L17_full.jpg} &
\includegraphics[width=0.108\textwidth]{figures/hero/cells/#1_L17_swa.jpg} &
\includegraphics[width=0.108\textwidth]{figures/hero/cells/#1_L17_wa2d.jpg} &
\rotatebox{90}{\scriptsize\;L17} \\
\end{tabular}%
}

\begin{figure*}[ht]
\centering
\subcaptionbox{\label{fig:bird_heatmap}}{\heatmapblock{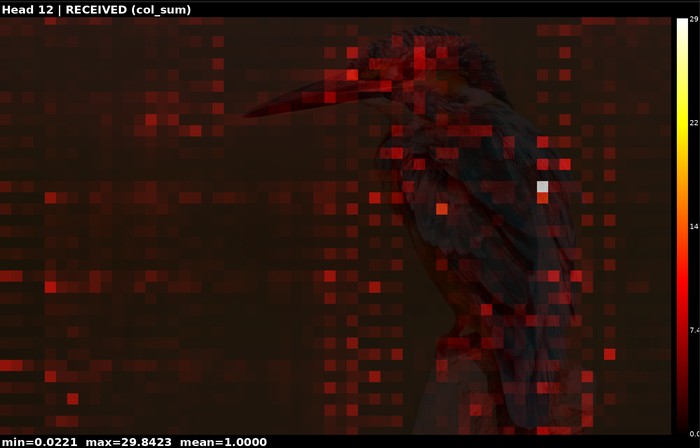}}\hfill
\subcaptionbox{\label{fig:twocats_heatmap}}{\heatmapblock{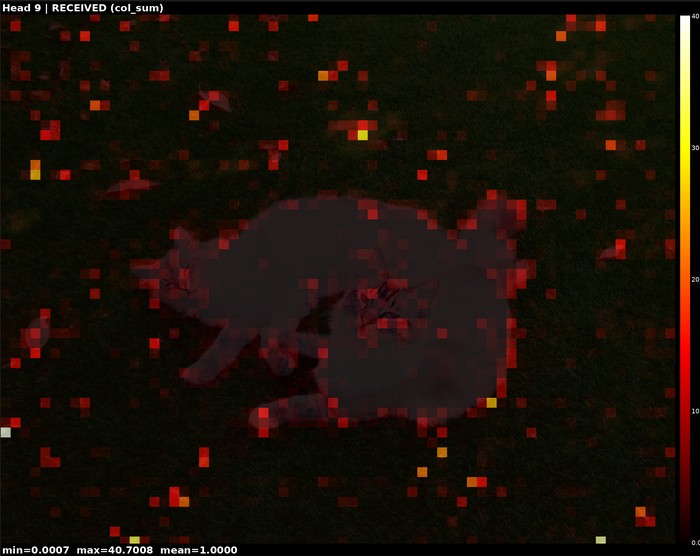}}\\[4pt]
\subcaptionbox{\label{fig:coco537964_heatmap}}{\heatmapblock{}}\hfill
\subcaptionbox{\label{fig:coco043816_heatmap}}{\heatmapblock{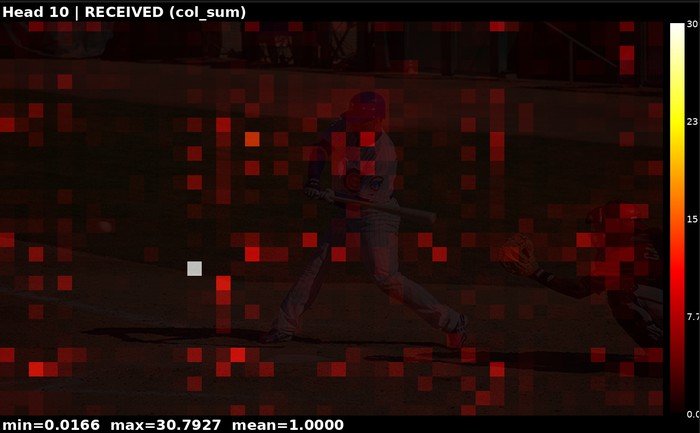}}
\caption{\textbf{Additional received-attention heatmaps} across four images, laid out as Figure~\ref{fig:hero}. Each block: first column = input image (top) and foreground mask (bottom); columns 2--4 = representative head heatmaps at L11 and L17 (layer labels on the right). Full Attention and Ariadne consistently show clear object/background differentiation; Chunk Window shows suppressed specialization across all images.}
\label{fig:appx_heatmaps}
\end{figure*}

\section{SHS Supplementary Data}
\label{sec:appx_shs_data}

\subsection{SHS-Index on Open-Source Visual Encoders and VLMs}

Table~\ref{tab:oss_shs} reports SHS-Index for 16 open-source visual encoders and VLMs spanning roughly five independent full-attention lineages, three training paradigms, diverse attention designs, and model sizes from 300M standalone ViTs to 72B-parameter VLMs. The additional standalone encoders are SigLIP2, DINOv2, and DINOv3 \citep{tschannen2025siglip2,oquab2023dinov2,simeoni2025dinov3}. SHS is computed on the same analysis images and foreground/background masks used for our controlled experiments. The heterogeneous models are used only to test the structural full-vs.-chunk separation, not for the benchmark correlation.

\begin{table*}[ht!]
\centering
\small
\setlength{\tabcolsep}{4pt}
\begin{tabular}{clllccc}
\toprule
\textbf{\#} & \textbf{Model} & \textbf{Family} & \textbf{Attention} & \textbf{ViT Spec} & \textbf{SHS-all} & \textbf{SHS-core} \\
\midrule
1 & InternVL3.5-38B & InternVL & Full & 6B (45L, 3200d, 25h) & \textbf{.648} & .627 \\
2 & InternVL3.5-8B & InternVL & Full & 300M (24L, 1024d, 16h) & .641 & \textbf{.651} \\
3 & Qwen3-VL-32B & Qwen3-VL & Full & SigLIP2-SO-400M & .633 & .644 \\
4 & InternVL3-8B & InternVL & Full & 300M (24L, 1024d, 16h) & .630 & .649 \\
5 & Qwen3-VL-8B & Qwen3-VL & Full & SigLIP2-SO-400M & .626 & .639 \\
6 & Qwen3-VL-4B & Qwen3-VL & Full & SigLIP2-Large (300M) & .624 & .639 \\
7 & Qwen3-VL-2B & Qwen3-VL & Full & SigLIP2-Large (300M) & .619 & .636 \\
8 & InternViT-300M & InternVL & Full & 300M (24L, 1024d, 16h) & .610 & .625 \\
9 & SigLIP2-so400m & SigLIP2 & Full & 400M & .638 & .646 \\
10 & DINOv2-L & DINOv2 & Full & ViT-L & .626 & .603 \\
11 & DINOv3-L & DINOv3 & Full & ViT-L & .642 & .627 \\
\midrule
12 & MiMo-VL-7B & MiMo & Chunk+Full & 675M (32L, 1280d, 16h) & .586 & .587 \\
13 & Qwen2.5-VL-3B & Qwen2.5-VL & Chunk+Full & 675M (32L, 1280d, 16h) & .586 & .586 \\
14 & Qwen2.5-VL-7B & Qwen2.5-VL & Chunk+Full & 675M (32L, 1280d, 16h) & .585 & .584 \\
15 & Qwen2.5-VL-72B & Qwen2.5-VL & Chunk+Full & 675M (32L, 1280d, 16h) & .585 & .585 \\
16 & Qwen2.5-VL-32B & Qwen2.5-VL & Chunk+Full & 675M (32L, 1280d, 16h) & .584 & .584 \\
\bottomrule
\end{tabular}
\caption{SHS-Index on 16 open-source visual encoders and VLMs. The full-attention group (mean SHS $\approx$ 0.631) and chunk-window group (mean SHS $=$ 0.585) show zero overlap: the lowest full-attention score is .610, above the highest chunk score .586. Qwen3-VL uses SigLIP2-SO-400M for its 8B/32B variants and SigLIP2-Large (300M) for its 2B/4B variants; Qwen2.5-VL uses an identical ViT across 3B--72B LLM sizes (SHS varies by ${<}\,0.002$). DINOv2 and DINOv3 were not trained with a language model, providing an additional check that the separation is not specific to VLM training.}
\label{tab:oss_shs}
\end{table*}

\subsection{Cross-domain robustness}
\label{sec:appx_cross_domain}

To test whether the SHS signal depends on the COCO analysis set, we recompute it on PASCAL VOC, OpenImages, and a chart/document-style image set. For charts, foreground consists of lines, bars, text, and axes identified by a simple grayscale threshold; light regions are treated as background. The full/chunk ordering is preserved in every domain, and the benchmark correlation remains significant. The Pearson values below are computed over all eight diagnostic configurations; representative rows are shown for readability.

\begin{table*}[ht!]
\centering
\small
\setlength{\tabcolsep}{6pt}
\begin{tabular}{@{}lcccc@{}}
\toprule
\textbf{Model} & \textbf{COCO} & \textbf{VOC} & \textbf{OpenImages} & \textbf{Chart} \\
\midrule
Full+Sink & .611 & .615 & .618 & .634 \\
Full & .606 & .609 & .614 & .627 \\
SWA+Sink (1D) & .598 & .596 & .608 & .625 \\
SWA+Sink (2D) & .597 & .600 & .598 & .626 \\
Chunk & .577 & .583 & .575 & .613 \\
\midrule
\textbf{Pearson $r$ (vs.\ 22-task)} & \textbf{.87} & \textbf{.80} & \textbf{.88} & \textbf{.75} \\
\textbf{$p$} & \textbf{.005} & \textbf{.016} & \textbf{.004} & \textbf{.031} \\
\bottomrule
\end{tabular}
\caption{Cross-domain SHS-Index robustness. Correlations use all eight diagnostic configurations at 8k; the five model rows are representative. The model ranking remains highly stable across domains (cross-domain rank Spearman $\rho{=}0.97$).}
\label{tab:cross_domain_shs}
\end{table*}

\subsection{Iteration-Level SHS Metrics}

\begin{figure}[ht!]
\centering
\includegraphics[width=\columnwidth]{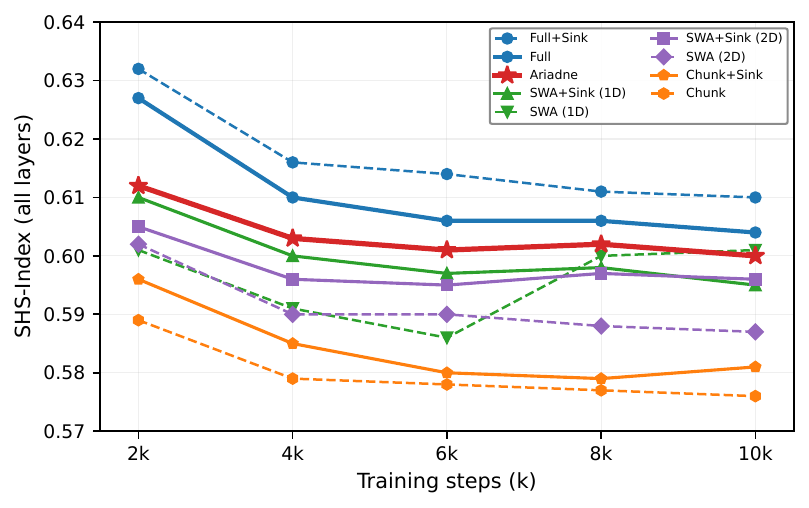}
\caption{SHS-Index across iterations. High/low separation stabilizes from 4k onward.}
\label{fig:shs_iter}
\end{figure}

\begin{figure}[ht!]
\centering
\includegraphics[width=\columnwidth]{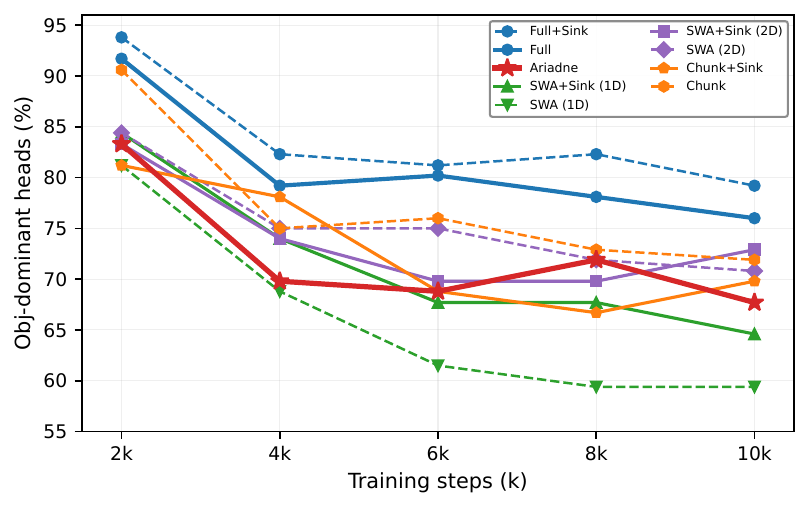}
\caption{Obj-dominant heads across iterations. SWA-1D without sink has the lowest percentage, while full attention remains highest.}
\label{fig:obj_pct_iter}
\end{figure}

This subsection provides iteration-level SHS metrics for all nine controlled architectures.
Figures~\ref{fig:shs_iter} and~\ref{fig:obj_pct_iter} visualize the trends from Tables~\ref{tab:shs_all_iter}, \ref{tab:shs_core_iter}, and~\ref{tab:obj_pct_iter}.
The absolute SHS-Index changes modestly over training, while the high/low group separation is already stable by 4k and changes by at most .003 from 8k to 10k for every model.
Obj-dominant percentage shows a similar downward trend, reflecting that heads increasingly split between object and background roles.

\subsection{SHS-Index Across Iterations (All Layers)}

Table~\ref{tab:shs_all_iter} reports the all-layer SHS-Index at each checkpoint.
The high/low group separation identified in Appendix~\ref{sec:appx_kmeans_robust} is visible from 4k onward; see Figure~\ref{fig:shs_iter} for a visualization.

\begin{table*}[t]
\centering
\small
\begin{tabular}{lccccccccc}
\toprule
\textbf{Iter} & \textbf{full\_s} & \textbf{full} & \textbf{4r3c} & \textbf{sink\_r} & \textbf{swa\_r} & \textbf{s2d\_s} & \textbf{swa2d} & \textbf{w\_s} & \textbf{wa2d} \\
\midrule
2k & .632 & .627 & .612 & .610 & .601 & .605 & .602 & .596 & .589 \\
4k & .616 & .610 & .603 & .600 & .591 & .596 & .590 & .585 & .579 \\
6k & .614 & .606 & .601 & .597 & .586 & .595 & .590 & .580 & .578 \\
8k & .611 & .606 & .602 & .598 & .600 & .597 & .588 & .579 & .577 \\
10k & .610 & .604 & .600 & .595 & .601 & .596 & .587 & .581 & .576 \\
\bottomrule
\end{tabular}
\caption{SHS-Index (mean over all 32 layers) across training iterations for all 9 controlled architectures. This is the primary aggregation used throughout the paper (Table~\ref{tab:three_metrics}, Pearson correlation in \S\ref{sec:validation}, and Figure~\ref{fig:hero_b}). Column headers abbreviated: full\_s = full\_sink, sink\_r = swa\_sink\_row, swa\_r = swa\_row, s2d\_s = swa2d\_sink, w\_s = wa2d\_sink.}
\label{tab:shs_all_iter}
\end{table*}

\subsection{SHS-Index Across Iterations (Core L9--21)}

Table~\ref{tab:shs_core_iter} restricts the computation to core layers L9--21, where head specialization is strongest. The gap between high and low groups widens compared to the all-layer average.

\begin{table*}[t]
\centering
\small
\begin{tabular}{lccccccccc}
\toprule
\textbf{Iter} & \textbf{full\_s} & \textbf{full} & \textbf{4r3c} & \textbf{sink\_r} & \textbf{swa\_r} & \textbf{s2d\_s} & \textbf{swa2d} & \textbf{w\_s} & \textbf{wa2d} \\
\midrule
2k & .651 & .644 & .612 & .615 & .604 & .605 & .606 & .597 & .586 \\
4k & .636 & .629 & .602 & .603 & .593 & .598 & .596 & .583 & .577 \\
6k & .635 & .624 & .604 & .600 & .588 & .597 & .599 & .578 & .576 \\
8k & .632 & .623 & .604 & .599 & .603 & .601 & .592 & .580 & .576 \\
10k & .630 & .622 & .605 & .594 & .601 & .598 & .593 & .580 & .574 \\
\bottomrule
\end{tabular}
\caption{SHS-Index (core layers L9--21) across training iterations.}
\label{tab:shs_core_iter}
\end{table*}

\subsection{Obj-Dominant Percentage Across Iterations}

Table~\ref{tab:obj_pct_iter} tracks the fraction of heads classified as object-dominant (majority vote across analysis images, core L9--21). Full attention models maintain the highest obj\% throughout; SWA 1D without sink develops the lowest, consistent with the sink-as-background-suppressor analysis in Appendix~\ref{sec:appx_mechanism}. See Figure~\ref{fig:obj_pct_iter} for a visualization.

\begin{table*}[t]
\centering
\small
\begin{tabular}{lccccccccc}
\toprule
\textbf{Iter} & \textbf{full\_s} & \textbf{full} & \textbf{4r3c} & \textbf{sink\_r} & \textbf{swa\_r} & \textbf{s2d\_s} & \textbf{swa2d} & \textbf{w\_s} & \textbf{wa2d} \\
\midrule
2k & 93.8 & 91.7 & 83.3 & 84.4 & 81.2 & 83.3 & 84.4 & 81.2 & 90.6 \\
4k & 82.3 & 79.2 & 69.8 & 74.0 & 68.8 & 74.0 & 75.0 & 78.1 & 75.0 \\
6k & 81.2 & 80.2 & 68.8 & 67.7 & 61.5 & 69.8 & 75.0 & 68.8 & 76.0 \\
8k & 82.3 & 78.1 & 71.9 & 67.7 & 59.4 & 69.8 & 71.9 & 66.7 & 72.9 \\
10k & 79.2 & 76.0 & 67.7 & 64.6 & 59.4 & 72.9 & 70.8 & 69.8 & 71.9 \\
\bottomrule
\end{tabular}
\caption{Percentage of obj-dominant heads (core L9--21, majority vote across analysis images).}
\label{tab:obj_pct_iter}
\end{table*}

\subsection{Mean rcv\_auroc\_obj Across Iterations}

Table~\ref{tab:auroc_obj_iter} reports the mean per-head AUROC for the object class (before the direction-agnostic $\max$ operation used in SHS-Index). This raw metric confirms that full attention models develop the strongest foreground discrimination, while chunk-window models remain the weakest at every checkpoint.

\begin{table*}[t]
\centering
\small
\begin{tabular}{lccccccccc}
\toprule
\textbf{Iter} & \textbf{full\_s} & \textbf{full} & \textbf{4r3c} & \textbf{sink\_r} & \textbf{swa\_r} & \textbf{s2d\_s} & \textbf{swa2d} & \textbf{w\_s} & \textbf{wa2d} \\
\midrule
2k & .618 & .608 & .566 & .576 & .552 & .567 & .570 & .566 & .557 \\
4k & .596 & .584 & .546 & .554 & .533 & .547 & .548 & .545 & .543 \\
6k & .589 & .574 & .545 & .540 & .524 & .539 & .546 & .529 & .537 \\
8k & .587 & .574 & .545 & .543 & .522 & .540 & .539 & .531 & .534 \\
10k & .581 & .566 & .543 & .532 & .519 & .537 & .537 & .530 & .531 \\
\bottomrule
\end{tabular}
\caption{Mean rcv\_auroc\_obj (core L9--21) across training iterations.}
\label{tab:auroc_obj_iter}
\end{table*}

\section{Mechanism Analysis Details}
\label{sec:appx_mechanism}

\subsection{SHS vs.\ Entropy vs.\ Distance}

\paragraph{Definitions.}
\emph{Attention entropy}: $H = -\sum_j a_j \log a_j$, measuring how peaked or diffuse a head's attention distribution is.
\emph{Attention distance}: the attention-weighted mean spatial distance between each query and its attended keys, measuring how far each token ``looks.''
Table~\ref{tab:three_metrics} reports all three metrics alongside benchmark for all 9 models at the primary checkpoint.

\begin{table*}[t]
\centering
\small
\begin{tabular}{llcccccccc}
\toprule
\textbf{Model} & \textbf{Layout} & \textbf{Attn} & \textbf{Sink} & \textbf{SHS} & \textbf{Entropy} & \textbf{Distance} & \textbf{Bench} \\
\midrule
full & N/A & Full & No & .606 & 4.584 & 14.922 & 40.92 \\
full\_sink & N/A & Full & Yes & .611 & 4.350 & 14.824 & 39.95 \\
4r3c & 1D rc & SWA & Yes & .602 & 2.768 & 13.950 & 40.40 \\
swa\_sink\_row & 1D r & SWA & Yes & .598 & 2.969 & 12.002 & 40.14 \\
swa\_row & 1D r & SWA & No & .600 & 3.010 & 11.713 & 38.80 \\
swa2d\_sink & 2D & SWA & Yes & .597 & 2.975 & 12.802 & 39.13 \\
swa2d & 2D & SWA & No & .588 & 2.914 & 11.802 & 38.09 \\
wa2d\_sink & 2D & Chunk & Yes & .579 & 2.428 & 11.146 & 37.94 \\
wa2d & 2D & Chunk & No & .577 & 2.476 & 10.690 & 37.35 \\
\bottomrule
\end{tabular}
\caption{SHS, Entropy, and Distance for all 9 models at the primary checkpoint. SHS-Index is the all-32-layer mean (matching Table~\ref{tab:shs_all_iter}; same convention used for the Pearson correlation in \S\ref{sec:validation}). Entropy and Distance are computed only over the 28 non-full-attn layers, since the 4 full-attn layers are identical across all 9 architectures and would dominate these distance/entropy averages.}
\label{tab:three_metrics}
\end{table*}

\paragraph{Cross-model Spearman correlation with benchmark (N=8, excl.\ full\_sink).}
Table~\ref{tab:metric_corr} compares the three metrics' correlation with benchmark performance.

\begin{table}[ht!]
\centering
\small
\begin{tabular}{lcc}
\toprule
\textbf{Metric} & $\boldsymbol{\rho}$ & $\boldsymbol{p}$ \\
\midrule
\textbf{SHS} & \textbf{+0.857} & \textbf{0.007} \\
Distance & +0.833 & 0.010 \\
Entropy & +0.524 & 0.183 \\
\bottomrule
\end{tabular}
\caption{Cross-model Spearman $\rho$ vs.\ benchmark (N=8). SHS is the strongest correlate; Entropy is not significant.}
\label{tab:metric_corr}
\end{table}

\paragraph{Pairwise deltas.}
Table~\ref{tab:pairwise_delta} isolates each architectural change to show which metrics track benchmark improvements.

\begin{table}[ht!]
\centering
\small
\setlength{\tabcolsep}{4pt}
\begin{tabular}{@{}lcccc@{}}
\toprule
\textbf{Change} & $\Delta$\textbf{SHS} & $\Delta$\textbf{Ent.} & $\Delta$\textbf{Dist.} & $\Delta$\textbf{Bench} \\
\midrule
+Sink (Chunk 2D) & +.002 & $-0.05$ & $+0.46$ & $+0.59$ \\
+Sink (SWA 2D) & +.009 & $+0.06$ & $+1.00$ & $+1.04$ \\
+Sink (SWA 1D) & $-.002$ & $-0.04$ & $+0.29$ & $+1.34$ \\
+Sink (Full) & +.005 & $-0.23$ & $-0.10$ & $-0.97$ \\
Chunk$\to$SWA (2D) & +.011 & $+0.44$ & $+1.11$ & $+0.74$ \\
Chunk$\to$SWA (2D+s) & +.018 & $+0.55$ & $+1.66$ & $+1.78$ \\
\textbf{2D$\to$1D (SWA)} & \textbf{+.012} & \textbf{+0.10} & \textbf{$-0.09$} & \textbf{+0.71} \\
2D$\to$1D (SWA+s) & +.001 & $-0.01$ & $-0.80$ & $+1.01$ \\
+RowCol & +.004 & $-0.20$ & $+1.95$ & $+0.26$ \\
\bottomrule
\end{tabular}
\caption{Pairwise deltas for each architectural change at the corrected 8k checkpoint. The 2D$\to$1D transition improves the benchmark, but individual local changes do not always move SHS and benchmark in the same direction; the relationship is a global cross-architecture pattern rather than a monotonic rule for every intervention. ($\Delta$SHS uses all-32-layer mean as in Table~\ref{tab:three_metrics}; $\Delta$Entropy/Distance use the non-full-attn layers.)}
\label{tab:pairwise_delta}
\end{table}

\subsection{Sink 2$\times$2 Analysis}

We analyze how sink bias (related to the attention sink phenomenon \citep{xiao2024streamingllm} and register tokens \citep{darcet2024vision}) differentially absorbs attention from background vs.\ object tokens across 5 sink-equipped models (Table~\ref{tab:sink_2x2}). The sink steal ratio is defined as: $\text{steal}(\text{cat}) = \text{absorbed\_sum}(\text{cat}) / (\text{rcv\_sum}(\text{cat}) + \text{absorbed\_sum}(\text{cat}))$, computed over core layers L9--L21 at the primary checkpoint.

\begin{table}[ht!]
\centering
\small
\setlength{\tabcolsep}{4pt}
\begin{tabular}{@{}lcccc@{}}
\toprule
\textbf{Model} & \textbf{bg stl.} & \textbf{obj stl.} & $\Delta$ & \textbf{bg$>$obj} \\
\midrule
full\_sink & 0.38\% & 0.23\% & +0.14\% & 97.9\% \\
swa\_sink\_row & 4.61\% & 2.86\% & +1.74\% & 96.9\% \\
swa\_sink\_4r3c & 3.79\% & 2.52\% & +1.26\% & 93.8\% \\
swa2d\_sink & 2.84\% & 1.68\% & +1.16\% & 96.9\% \\
wa2d\_sink & 4.32\% & 3.69\% & +0.64\% & 78.1\% \\
\bottomrule
\end{tabular}
\caption{Sink steal analysis (all heads, core L9--21 at the primary checkpoint). bg/obj stl.\ = fraction of received attention redirected to sink. In SWA models, 94--97\% of heads steal more from bg than obj, i.e., sink acts as a \emph{background suppressor}. In wa2d\_sink, directionality drops to 78\%, explaining why chunk+sink fails.}
\label{tab:sink_2x2}
\end{table}

\subsection{2D Incompatibility Hypothesis}

Under 2D block layout, the sliding window's left and right neighbors in 1D sequence map to spatially disjoint regions of the image at every window boundary. This forces each head to attend to a mixture of spatially near and spatially distant tokens within the same window, diluting the spatial coherence that enables meaningful object/background discrimination.

Evidence (Table~\ref{tab:pairwise_delta}): the 2D$\to$1D transition yields the largest single-step benchmark improvement (+1.94 without sink, +1.22 with sink), yet Entropy and Distance are essentially unchanged ($\Delta$Entropy $= +0.10$, $\Delta$Distance $= -0.09$). Only SHS increases (+0.013), suggesting that the improvement comes from spatial coherence enabling more meaningful head specialization, not from changes in attention sharpness or receptive field size.

\section{SHS-Benchmark Correlation Details}
\label{sec:appx_correlation}

\subsection{Statistical Methods}

Our primary analysis is a Pearson correlation between SHS-Index and the 22-task benchmark average across the eight diagnostic configurations (\S\ref{sec:validation}). We report the correlation both \emph{before} introducing Ariadne (the eight diagnostic configurations alone) and \emph{after} including it (all nine controlled architectures), at the primary and final checkpoints, and we also report the 20-image-task counterpart for completeness. As a robustness check, we additionally apply $K$-means clustering ($k{=}2$) on SHS-Index followed by a Mann-Whitney $U$\,test on the resulting groups (\S\ref{sec:appx_kmeans_robust}); the two analyses agree on the qualitative conclusion.

\subsection{Robustness: K-means Clustering and Mann-Whitney $U$ Test}
\label{sec:appx_kmeans_robust}

As a non-parametric robustness check, $K$-means ($k{=}2$) on SHS-Index automatically separates the eight diagnostic configurations into a high-SHS group (5 models: full, full\_sink, swa\_sink\_row, swa\_row, swa2d\_sink; SHS $\geq 0.597$) and a low-SHS group (3 models: swa2d, wa2d\_sink, wa2d; SHS $\leq 0.588$). Adding Ariadne keeps the low-SHS group unchanged and expands the high-SHS group to 6 models. In both settings, the groups show \emph{zero benchmark overlap} at the corrected primary checkpoint: the lowest-scoring high-SHS model (38.80) exceeds the highest-scoring low-SHS model (38.09). A Mann-Whitney $U$\,test on the 20-image-task average confirms significance (Table~\ref{tab:shs_kmeans}).

\begin{table}[ht!]
\centering
\small
\setlength{\tabcolsep}{3pt}
\begin{tabular}{@{}lcccc@{}}
\toprule
\textbf{Setting} & \textbf{High} & \textbf{Low} & $p$ (U) & $d$ \\
\midrule
8 diagnostic @8k & 38.80--40.92 & 37.35--38.09 & \textbf{.018} & 3.65 \\
9 all @8k & 38.80--40.92 & 37.35--38.09 & \textbf{.012} & \textbf{3.91} \\
9 all @10k & 38.01--40.40 & 35.21--38.27 & \textbf{.024} & 2.37 \\
\bottomrule
\end{tabular}
\caption{Robustness check: $K$-means ($k{=}2$) on SHS-Index partitions the controlled configurations into high-SHS and low-SHS groups (benchmark ranges shown). Mann-Whitney $U$\,test on the 20-image-task average confirms significant separation. Cohen's $d$ uses unweighted pooled SD. The 9-model grouping remains significant at 10k.}
\label{tab:shs_kmeans}
\end{table}

\section{SHS Collective Property Experiments}
\label{sec:appx_collective}

\subsection{Head Elimination}

We test whether individual high-SHS or low-SHS heads are more important by replacing specified heads' V outputs with their calibration-set mean (mean ablation), following the head importance analysis methodology of \citet{michel2019sixteen}. For each model, we select the 8 heads with highest and 8 with lowest core-layer AUROC, eliminate them separately, and compare the benchmark drop.

\paragraph{High vs.\ low SHS heads and all-16 elimination.}
Table~\ref{tab:head_elim_hilo} combines both experiments on the full model. The upper panel shows that eliminating 8 high-SHS heads causes the same benchmark drop as eliminating 8 low-SHS heads (avg $\Delta {=} {-}0.18$); individual heads are interchangeable regardless of specialization direction. The lower panel shows that eliminating all 16 core-layer heads simultaneously causes dramatic drops on vision-dependent tasks, confirming that SHS is a collective property.

\begin{table}[ht!]
\centering
\small
\setlength{\tabcolsep}{4pt}
\begin{tabular}{@{}lrr@{}}
\toprule
\multicolumn{3}{@{}l}{\textit{Panel A: High-8 vs.\ Low-8 (avg over 23 tasks)}} \\
\midrule
\textbf{Condition} & \textbf{$\Delta$ vs.\ base} & \\
\cmidrule(r){1-2}
Remove 8 hi-SHS & $-1.25$ & \\
Remove 8 lo-SHS & $-1.07$ & \\
hi $-$ lo & $-0.18$ & (14/23 tasks hi$<$lo) \\
\midrule
\multicolumn{3}{@{}l}{\textit{Panel B: All 16 heads removed}} \\
\midrule
\textbf{Task} & \textbf{Base} & $\boldsymbol{\Delta}$ \\
\cmidrule(r){1-3}
V* & 48.69 & $-13.1$ \\
OCRBench & 68.00 & $-16.7$ \\
DocVQA & 53.46 & $-13.8$ \\
MMMU & 29.11 & $+0.1$ \\
\bottomrule
\end{tabular}
\caption{Head elimination on the full model (core layers). \textbf{Panel A}: removing 8 high-SHS vs.\ 8 low-SHS heads yields negligible difference; individual heads are interchangeable. \textbf{Panel B}: removing all 16 heads collapses vision-dependent tasks; MMMU (reasoning-heavy) is unaffected. The collective specialization pattern is indispensable. The head-elimination suite (23 tasks) is a superset of the 22-task primary benchmark, with sub-scores (e.g., MME-cognition vs.\ MME-perception) reported separately for finer-grained per-task analysis.}
\label{tab:head_elim_hilo}
\end{table}

\paragraph{Cross-model consistency.}
Table~\ref{tab:head_elim_cross} extends the high-vs-low analysis to three architectures.

\begin{table}[ht!]
\centering
\small
\begin{tabular}{lcc}
\toprule
\textbf{Model} & \textbf{Tasks} & \textbf{avg (hi $-$ lo)} \\
\midrule
full & 23 & $-0.18$ \\
wa2d & 23 & $+0.20$ \\
swa\_sink\_row & 25 & $-0.27$ \\
\bottomrule
\end{tabular}
\caption{High-low SHS head elimination across models. No model shows a significant difference. Task counts vary slightly (23--25) across models because evaluation suites differ in which sub-scores are reported separately at this checkpoint; the conclusion (no high/low gap) is robust to this choice.}
\label{tab:head_elim_cross}
\end{table}

\subsection{QK Weight Norm vs.\ SHS}

We compute the per-head Frobenius norm of Q and K weight matrices and correlate with per-head SHS (AUROC). Table~\ref{tab:qk_norm} shows Spearman $\rho$ between QK norm product and per-head AUROC for each model.

\begin{table}[ht!]
\centering
\small
\setlength{\tabcolsep}{3.5pt}
\begin{tabular}{@{}lccccc@{}}
\toprule
& Q & K & V & O & Q$\!{\times}\!$K \\
\midrule
full & +.09 & +.11 & $-.06$ & +.05 & +.10 \\
4r3c & +.15 & +.16 & +.01 & +.05 & +.16 \\
sink\_row & +.00 & +.02 & $-.03$ & +.13 & +.01 \\
swa\_row & +.14 & +.14 & +.06 & +.16 & +.14 \\
s2d\_sink & +.33 & +.32 & +.05 & +.19 & +.33 \\
swa2d & +.19 & +.20 & +.01 & +.12 & +.20 \\
w\_sink & +.12 & +.12 & $-.10$ & +.07 & +.12 \\
wa2d & +.27 & +.28 & $-.01$ & +.14 & +.28 \\
\midrule
\textbf{Mean} & \textbf{+.16} & \textbf{+.17} & \textbf{$-.01$} & \textbf{+.11} & \textbf{+.17} \\
\bottomrule
\end{tabular}
\caption{Spearman $\rho$ between per-head weight Frobenius norms and SHS. Q/K show weak positive correlations (8/8 positive); V shows none (mean $-.01$).}
\label{tab:qk_norm}
\end{table}

\subsection{QK Equalization Experiment}

We artificially equalize all heads' QK norms to the per-layer maximum, then recompute SHS. If QK norm causes SHS, equalization should increase SHS. Instead, SHS drops in all 5 models tested (Table~\ref{tab:qk_eq}):

\begin{table}[ht!]
\centering
\small
\begin{tabular}{lccc}
\toprule
\textbf{Model} & \textbf{Base SHS} & \textbf{Eq.\ SHS} & $\boldsymbol{\Delta}$ \\
\midrule
full & .6237 & .6178 & $-0.0059$ \\
4r3c & .6044 & .6001 & $-0.0044$ \\
swa\_sink\_row & .5993 & .5958 & $-0.0035$ \\
swa\_row & .6036 & .5997 & $-0.0038$ \\
wa2d & .5763 & .5751 & $-0.0013$ \\
\bottomrule
\end{tabular}
\caption{QK equalization: SHS drops in all 5 models. The positive QK-SHS correlation is a consequence, not a cause, of head specialization.}
\label{tab:qk_eq}
\end{table}

\section{Efficiency Analysis}
\label{sec:appx_efficiency}

Table~\ref{tab:efficiency_full} reports ViT forward pass timing across six representative resolutions.
All timings measured on NVIDIA Hopper GPU, FlashAttention-3 \citep{shah2024flashattention3} with \texttt{s\_aux=yes}, bfloat16, 4 images batched (varlen contiguous batching), warmup 10, iterations 10. ViT: 32 layers, hidden 1280, 16 heads, head\_dim 80, MLP 3456. Full-attention layers at positions [7, 15, 23, 31].
TFLOP counts theoretical attention FLOPs ($QK^\top$ + softmax$\cdot V$, multiply-add = 2 ops).
\texttt{attn\_op} is the FlashAttention kernel only (no QKV/proj linear; wa2d includes reorder overhead).
\texttt{e2e} is the full transformer block (norm + QKV + RoPE + attn + proj + norm + MLP).

\begin{table*}[ht!]
\centering
\small
\setlength{\tabcolsep}{3.5pt}
\begin{tabular}{lcrrrrrrrrrrrrr}
\toprule
& & \multicolumn{3}{c}{\textbf{Full}} & \multicolumn{3}{c}{\textbf{wa2d (Chunk)}} & \multicolumn{3}{c}{\textbf{SWA-row}} & \multicolumn{3}{c}{\textbf{SWA-sink-row}} \\
\cmidrule(lr){3-5} \cmidrule(lr){6-8} \cmidrule(lr){9-11} \cmidrule(lr){12-14}
\textbf{Res} & \textbf{Tok} & \textbf{FLOP} & \textbf{op} & \textbf{e2e} & \textbf{FLOP} & \textbf{op} & \textbf{e2e} & \textbf{FLOP} & \textbf{op} & \textbf{e2e} & \textbf{FLOP} & \textbf{op} & \textbf{e2e} \\
\midrule
$224^2$ & 256 & 0.04 & 1.2 & 22.1 & 0.01 & 2.8 & 24.4 & 0.02 & 1.2 & 22.1 & 0.02 & 1.2 & 22.3 \\
$448^2$ & 1024 & 0.69 & 2.2 & 32.1 & 0.12 & 4.5 & 32.9 & 0.16 & 1.4 & 31.4 & 0.16 & 1.4 & 31.3 \\
$672^2$ & 2304 & 3.48 & 7.6 & 68.7 & 0.52 & 9.7 & 68.4 & 0.60 & 2.9 & 64.6 & 0.60 & 2.9 & 64.6 \\
$\mathbf{896^2}$ & \textbf{4096} & \textbf{11.0} & \textbf{23.9} & \textbf{124.4} & 1.52 & 17.5 & 115.3 & 1.68 & 6.2 & 109.0 & \textbf{1.68} & \textbf{6.3} & \textbf{109.0} \\
$1120^2$ & 6400 & 26.8 & 55.2 & 211.2 & 3.59 & 28.8 & 178.1 & 3.83 & 12.1 & 168.5 & 3.83 & 12.0 & 168.6 \\
$\mathbf{1344^2}$ & \textbf{9216} & \textbf{55.7} & \textbf{110.3} & \textbf{330.0} & 7.30 & 44.3 & 255.5 & 7.64 & 21.2 & 242.2 & \textbf{7.64} & \textbf{21.1} & \textbf{242.9} \\
\bottomrule
\end{tabular}
\caption{Four-operator ViT profiling at representative resolutions. FLOP = attention TFLOP; op = FlashAttention kernel time (ms); e2e = full block time (ms). SWA-sink-row $\equiv$ SWA-row: sink adds \emph{zero} overhead in this run (FA3 native \texttt{s\_aux} support). The updated Full--Ariadne scaling sweep is reported in Table~\ref{tab:efficiency_scaling}.}
\label{tab:efficiency_full}
\end{table*}

\begin{table*}[ht!]
\centering
\small
\setlength{\tabcolsep}{5pt}
\begin{tabular}{@{}lrrrrrr@{}}
\toprule
\textbf{Resolution} & \textbf{Tokens} & \textbf{Full attn.} & \textbf{Ariadne attn.} & \textbf{Full e2e} & \textbf{Ariadne e2e} & \textbf{e2e saving} \\
\midrule
$448^2$ & 1{,}024 & 0.69T & 0.16T & 31.3 & 30.6 & $-2.2\%$ \\
$672^2$ & 2{,}304 & 3.48T & 0.60T & 66.9 & 62.8 & $-6.1\%$ \\
$896^2$ & 4{,}096 & 11.0T & 1.68T & 123.5 & 106.8 & $-13.5\%$ \\
$1120^2$ & 6{,}400 & 26.8T & 3.83T & 210.1 & 166.1 & $-20.9\%$ \\
$1344^2$ & 9{,}216 & 55.7T & 7.64T & 327.5 & 240.0 & $-26.7\%$ \\
$1568^2$ & 12{,}544 & 103.1T & 13.8T & 500.2 & 331.0 & $-33.8\%$ \\
$1792^2$ & 16{,}384 & 175.9T & 23.2T & 726.2 & 440.0 & $-39.4\%$ \\
\bottomrule
\end{tabular}
\caption{Updated Full--Ariadne efficiency sweep on NVIDIA Hopper with FlashAttention-3. Attention columns report theoretical attention FLOPs; e2e columns report complete ViT block time in milliseconds. The end-to-end saving grows with resolution, which is the regime targeted by hybrid attention.}
\label{tab:efficiency_scaling}
\end{table*}

\paragraph{Key conclusions.}
\begin{itemize}
\item \textbf{Sink has zero overhead}: SWA-sink-row and SWA-row produce identical \texttt{attn\_op} and e2e timings at every resolution, confirming that FlashAttention-3's \citep{shah2024flashattention3} native \texttt{s\_aux} parameter incurs no additional computation.
\item \textbf{SWA faster than chunk}: Despite a larger effective window (128 vs.\ 64 tokens), SWA's kernel time is dramatically lower than wa2d (6.3\,ms vs.\ 17.5\,ms at $896^2$). FlashAttention-3's contiguous SWA kernel is better optimized than the varlen interface used for chunk windows \citep{dao2022flashattention, dao2023flashattention2}. The wa2d kernel time also includes the overhead of 2D window reordering.
\item \textbf{Savings grow with resolution}: In the updated sweep, Ariadne saves 13.5\% e2e at $896^2$ and 39.4\% at $1792^2$. Full attention FLOPs scale quadratically, while the local component of Ariadne scales approximately linearly.
\end{itemize}

\section{Why Sink Bias Hurts Full Attention}
\label{sec:appx_full_sink}

The Full+Sink model has the highest SHS-Index (0.611) yet ranks below Full on the corrected benchmark (39.95 vs.\ 40.92), a $-0.97$ drop.
In contrast, sink bias consistently \emph{improves} SWA models ($+0.5$ to $+2.1$).
Three lines of evidence explain this asymmetry.

\paragraph{Full attention already has implicit sinks.}
\citet{guo2024active} show that full-attention heads naturally develop an \emph{active-dormant} mechanism: for each input, some heads route attention to implicit sink tokens whose value-state norms approach zero, effectively implementing a learned, input-dependent no-op.
\citet{barbero2025why} further demonstrate that sinks function as anti-over-mixing regulators, and that larger context windows develop stronger implicit sinks.
A learnable sink bias $s^{(h)}_{\mathrm{aux}}$ is a fixed, input-independent scalar; it competes with the pre-existing input-dependent mechanism rather than complementing it.

\paragraph{SWA lacks this implicit mechanism.}
With only 128 tokens per window, SWA models cannot develop effective implicit sinks; the local context is too limited.
The softmax constraint forces each query to distribute its full attention budget within the window, wasting mass on uninformative background tokens.
Here, an explicit sink bias fills a genuine gap: it provides the no-op channel that the architecture cannot learn on its own.

\paragraph{Over-specialization reduces diversity.}
Full+Sink's obj-dominant head fraction is 82.3\% (the highest among all 9 models), meaning sink bias pushes too many heads toward foreground.
This reduces the collective role diversity that \S\ref{sec:appx_collective} identifies as indispensable, explaining why SHS-Index rises (heads are individually more specialized) while benchmark falls (the ensemble loses complementary background expertise).

This analysis is consistent with \citet{gu2025attention}, who show that attention sinks arise from the softmax normalization constraint and function as key biases rather than informative signals; adding a second explicit bias atop an already-effective implicit one yields diminishing or negative returns.

\section{Pearson Correlation}
\label{sec:appx_pearson}

Table~\ref{tab:shs_pearson} reports the full Pearson $r$ matrix. The relationship is strong at the corrected primary checkpoint under both task averages (20 image and 22 with video) and under both $N{=}8$ (eight diagnostic configurations) and $N{=}9$ (adding Ariadne). At the final checkpoint the correlation weakens; those 10k rows retain the original MMBench-EN evaluation because the re-evaluation was performed at 8k only.

\begin{table}[ht!]
\centering
\small
\setlength{\tabcolsep}{4pt}
\begin{tabular}{@{}llcccc@{}}
\toprule
\textbf{Step} & \textbf{Setting} & \multicolumn{2}{c}{\textbf{22-task}} & \multicolumn{2}{c}{\textbf{20 image-task}} \\
\cmidrule(lr){3-4}\cmidrule(l){5-6}
& & $r$ & $p$ & $r$ & $p$ \\
\midrule
8k & $N{=}8$ (diagnostic) & \textbf{.858} & \textbf{.006} & .877 & .004 \\
8k & $N{=}9$ (incl.\ Ariadne) & \textbf{.853} & \textbf{.003} & .875 & .002 \\
\midrule
10k & $N{=}8$ & .66 & .077 & .67 & .067 \\
10k & $N{=}9$ & .67 & .048 & .69 & .041 \\
\bottomrule
\end{tabular}
\caption{Pearson correlation between SHS-Index and benchmark across controlled architectures. The 8k rows use corrected MMBench-EN scores; bootstrap 95\% CI for the primary 22-task setting is $[0.51,0.98]$, and an exact two-sided permutation test gives $p{=}0.008$. The 10k rows retain the original evaluation and are included as a sensitivity analysis.}
\label{tab:shs_pearson}
\end{table}

\subsection{Bootstrap and exact permutation tests}

Because the controlled comparison has a small number of configurations, we complement Pearson's parametric $p$-value with bootstrap intervals and exact permutation tests. The primary 22-task correlation has a 95\% bootstrap interval whose lower endpoint remains above .50; the exact test enumerates all label permutations and makes no distributional assumption.

\begin{table}[ht!]
\centering
\small
\setlength{\tabcolsep}{3.5pt}
\begin{tabular}{@{}lccc@{}}
\toprule
\textbf{Setting} & $r$ & \textbf{Bootstrap 95\% CI} & \textbf{Exact $p$} \\
\midrule
$N{=}8$, 22-task & .858 & [.51, .98] & .008 \\
$N{=}8$, 20-image & .877 & [.63, .98] & .006 \\
$N{=}9$, 22-task & .853 & [.50, .98] & .004 \\
$N{=}9$, 20-image & .875 & [.63, .98] & .002 \\
\bottomrule
\end{tabular}
\caption{Bootstrap intervals (10k resamples) and exact two-sided permutation $p$-values computed from the corrected 8k benchmark scores. The $N{=}8$ permutation enumerates all 40,320 assignments.}
\label{tab:shs_bootstrap}
\end{table}

\end{document}